%% file: main.tex
\documentclass[10pt,twocolumn,letterpaper]{article}

\usepackage{cvpr}              


\definecolor{cvprblue}{rgb}{0.21,0.49,0.74}
\usepackage[pagebackref,breaklinks,colorlinks,allcolors=cvprblue]{hyperref}

\def\paperID{10083} 
\def\confName{CVPR}
\def\confYear{2025}

\title{STUPD: A Synthetic Dataset for Spatial and Temporal Relation Reasoning}

\author{Palaash Agrawal\\
Center for Frontier AI Research\\
Agency for Science, Technology, and Research, Singapore \\
{\tt\small agrawal\_palaash@cfar.a-star.edu.sg}\\
\and
Haidi Azaman\\
National University of Singapore\\
{\tt\small haidiazaman@gmail.com}\\
\and
Cheston Tan\\
Center for Frontier AI Research\\
Agency for Science, Technology, and Research, Singapore \\
{\tt\small cheston-tan@i2r.a-star.edu.sg}}

\begin{document}
\maketitle
\input{content/0.abstract}

\section{Introduction}
\label{sec:intro}
\input{content/1.introduction}

\section{Related Work}
\label{sec:related_work}

\input{content/2.related_work}

\section{The STUPD Dataset}
\label{sec:dataset}
\input{content/3.dataset}

\section{Baselines}
\label{sec:baselines}
\input{content/4.baselines}

\section{Limitations and Future Work}
\label{sec:limitations}
\input{content/5.limitations}

\section{Conclusions}
\label{sec:baselines}
\input{content/6.conclusion}
{
    \small
    \bibliographystyle{ieeenat_fullname}
    \bibliography{main}
}

\clearpage
\appendix
\section{APPENDIX}
\input{content/Appendix}

\end{document}

%% file: content/0.abstract.tex
\begin{abstract}
{Identifying relations between objects is crucial for understanding the semantics of a visual scene. It is also an essential step in order to bridge visual and language models. However, current state-of-the-art computer vision models still lack the ability to perform spatial reasoning well. Existing datasets mostly cover a relatively small number of spatial relations, all of which are static relations that do not intrinsically involve motion. In this paper, we propose the \textbf{S}patial and \textbf{T}emporal \textbf{U}nderstanding of \textbf{P}repositions \textbf{D}ataset (\textbf{STUPD}) -- a large-scale video dataset for understanding spatial and temporal relationships derived from prepositions of the English language. The dataset contains 150K visual depictions (videos and images), consisting of 30 static and dynamic spatial prepositions, in the form of object interaction simulations generated synthetically. In addition to spatial relations, we also propose 50K visual depictions across 10 temporal relations, consisting of videos depicting event/time-point interactions. To our knowledge, no dataset exists that represents temporal relations through visual settings. In this dataset, we also provide 3D information about object interactions such as frame-wise coordinates, and descriptions of the objects used. This synthetic dataset aims to help models perform better in visual relationship detection in real-world settings. We demonstrate an increase in the performance of various models over 2 real-world datasets (ImageNet-VidVRD and Spatial Senses) when pretrained on the spatial STUPD dataset, and over the Kinetics-400 dataset on the temporal STUPD dataset, in comparison to other pretraining datasets. 
}
\end{abstract}

%% file: content/1.introduction.tex
Identifying relationships between objects is crucial for semantic understanding of the visual world. However, current state-of-the-art computer vision models still struggle to understand relationships~\cite{Conwell2022testing, DallEval, Liu2021, Thrush2022winoground, Ramesh2022hierarchical, IETrans}. For instance, even for simple relations in 2D pixel space such as ``left'', ``right'', ``above'' and ``below'', Cho et al.~\cite{DallEval} found a large gap between upper-bound accuracy and the performance of generative transformers. 

In human languages, relational concepts are conveyed using prepositions, which are words used \textit{``to show a relationship in space or time''}~\cite{TPP}. Examples of prepositions include ``above'', ``before'' and ``with''. Existing computer vision datasets cover English parts-of-speech such as nouns/objects~\cite{Imagenet, CIFAR}, verbs/actions~\cite{Charades, ActionGenome, Kinetics}, adjectives/attributes~\cite{RelativeAttributes, VisualGenome}, etc. However, despite their importance, prepositions are significantly understudied in computer vision as a distinct class of concepts. 

Prepositions may have one or more senses, which are distinct definitions of a word in different contexts. For example, the preposition ``against'' has 2 distinct spatial senses~\cite{TPP}. One refers to a situation where 2 objects are moving in opposite directions and the other where an object is leaning on another. For simplicity, we will henceforth use the term ``preposition'' to refer to both prepositions (the words) and their senses (the definitions), except where clear distinctions are required. A detailed glossary of all terms introduced in this paper is included in the Appendix.


From Table~\ref{table:1}, it can be observed that image datasets that contain hundreds to thousands of relation classes actually have fewer than 30 prepositions (an exception is the recent VSR dataset~\cite{VSR} which covers 65 
prepositions).
As for existing video datasets, only 6-8 prepositions are covered. Furthermore, datasets thus far contain only \textit{static} prepositions, which are prepositions that do not necessarily involve any motion, such as ``above'' and ``behind''. The vast majority of such examples come from very simple and intuitive preposition classes such as ``on'' or ``near'', which are easier to label by human annotators.  None of the existing datasets include \textit{dynamic} prepositions, which are prepositions that intrinsically involve motion, such as ``into'', ``onto'', etc. Finally, existing datasets are also extremely imbalanced due to the long-tailed distribution of relationship occurrences. 

This kind of highly restrictive relational domain in existing datasets is not an effective approach towards visual reasoning, because it only focuses on position, while ignoring many fundamental relational characteristics, such as relative speed, contact and physical forces of interaction. The prospect of the ability to distinguish between different spatial (as well as temporal) configurations with higher granularity, thus, makes it worthwhile to study the wider variety of prepositions for effective visual reasoning. Through this, datasets can be richer in information, and models would be able to differentiate between many related but different relational categories (such as ``above'' and ``over''). A granular understanding of prepositional relations also allows for better understanding of language semantics, which is an equally important and complementary aspect of visual reasoning in the understanding of a scene.

Apart from spatial reasoning, understanding temporal relations is also crucial for visual reasoning. Many relations require understanding dynamics of interactions over time. Visual representation of temporal relationships is challenging because temporal concepts are unintuitive to visualize. This is one of the reasons why temporal relations are heavily underrepresented in visual reasoning datasets. Without effectively understanding temporal relations, spatial relations remain isolated, and their progression cannot be understood. Thus spatial and temporal relations should be treated as equally important aspects of visual reasoning.


\textbf{Contributions.} To address these issues, we created the Spatial and Temporal Understanding of Prepositions Dataset (STUPD) -- the first dataset to include dynamic spatial prepositions and temporal relations. The contributions of this paper are as follows:
\begin{enumerate}
    \item \textbf{Comprehensive synthetic dataset for spatial relations}: This paper introduces a dataset consisting of 150,000 images and videos that capture 30 different spatial relations. The dataset incorporates realistic physical interactions using a sophisticated physics engine coupled with diverse backgrounds. 
    \item \textbf{Comprehensive synthetic dataset for temporal relations}: In addition to spatial relations, this paper introduces a separate dataset comprising 50,000 video-pairs depicting 10 different temporal relations. Through this, the paper also introduces a definitive framework for defining and distinguishing between different temporal relations, for future works to build on.
    \item \textbf{Detailed 3D meta-information}: To enhance the quality and usability of the dataset, each image and video in the dataset is accompanied by detailed 3D information and bounding box annotations.
    \item \textbf{Effective pre-training dataset with real-world applicability}: The proposed datasets are primarily designed to serve as a highly effective pre-training resource for computer vision models. Pre-training on this dataset provides a solid foundation for subsequent fine-tuning on real-world datasets. Later in the paper, we demonstrate that pretraining on STUPD increases performance on real-world visual reasoning tasks. 
\end{enumerate}

%% file: content/2.related_work.tex
 \begin{table}
 \centering
\begin{center}
\resizebox{1.1\linewidth}{!}{
\begin{tabular}{llllllll}
 \hline
  \textbf{Type} & \textbf{Dataset} & \textbf{\rotatebox[origin=c]{90}{ 3D Info? }} & \textbf{\rotatebox[origin=c]{90}{ \# Preps }} & \textbf{\rotatebox[origin=c]{90}{Dyn?}} & \textbf{\rotatebox[origin=c]{90}{Tem?}} & \begin{tabular}[c]{@{}l@{}}\textbf{Real/}\\\textbf{Synth}\end{tabular} & \textbf{Size}
  \\  [0.0ex] 
 \hline\hline

Image & VSR~\cite{VSR} (2022)                 & N &{\textbf{65}} & N & N & Real & 10K\\
 Image & Liu et al.~\cite{Liu2021} (2021)    & N &    6 & N & N & Synth & 83K \\
 Image & Rel3D~\cite{Rel3d} (2020)    & {\textbf{Y}} &   25 & N & N & Synth & 27.3K \\
 Image & SpatialSense~\cite{Spatialsense}   & N &    9 & N & N & Real & 11.5K\\
 & (2019) \\
 Image & CLEVR~\cite{CLEVR}(2017) & N &    4 & N & N & Synth & 100K\\
 Image & Visual Genome- & N &   21 & N & N &  Real  & 108K\\
 &  50~\cite{VG50} (2017) &  \\
 Image & VRD~\cite{VRD}(2016) & N &   24 & N & N &  Real  & 5K\\

Image & Scene Graphs~\cite{SceneGraphs}   & N &   29 & N & N &  Real & 5K\\
&  (2015)\\
Image & PTR ~\cite{hong2021ptr} (2021)   & Y &   4 & N & N &  Synth & 52K\\
 \hline

 VR& iGibson 2.0~\cite{iGibson2} & {\textbf{Y}} &    6 & N & N & Synth & N/A\\
 & (2021)\\
 
\hline

 Video & CATER~\cite{CATER} (2020) & N &    7 & N &  {\textbf{Y}} &  Real & 5.5K\\

 Video & Action Genome & N &    6 & N & N &  Real & 1.75K\\

 &~\cite{ActionGenome} (2020) \\

 Video & VidOR~\cite{VidOR} (2019) & N &    8 & N & N &  Real & 10K\\

 \hline
 Video & \textbf{STUPD (ours)} (2024) &  {\textbf{Y}} & {40} & {\textbf{Y}} & {\textbf{Y}}  & Synth & \textbf{200K}\\
 \end{tabular}
 }
\end{center}
\caption{Comparison of relations datasets. (Preps = Prepositions (relations), Dyn = Dynamic in nature, Tem = Temporal, Synth = Synthetically generated)}
\label{table:1}
\end{table}

\subsection{Vision Datasets}
In recent years, image-based datasets have attempted to present spatial relationships through simple 2D object interactions~\cite{Spatialsense, VisualGenome, VRD}. However, 2D interactions restrict the scope of distinguishable visual relations. Synthetic datasets have proved to be useful for many use cases, and are becoming increasingly popular as a way to bridge the information gap in image datasets through 3D spatial information~\cite{CLEVR, Liu2021}. An example is CLEVR~\cite{CLEVR}, which consists of synthetic images with objects arranged in various configurations to promote systematic spatial reasoning.

However, synthetic datasets in this domain do not provide three-dimensional information about object location or orientation, rendering the perceptual input provided as effectively two-dimensional. Some works such as \cite{Rel3d} provide annotated synthetic 3D scenes. This allows models to better understand object interactions and distinguish between subtle visual relations such as impact and contact. 


A common theme across different visual relation datasets is to compound prepositional relations with other complex actions~\cite{somethingsomething}. For instance, in the Action Genome dataset~\cite{ActionGenome}, the action ``sitting'' and the preposition ``on'' are combined into a single relation ``sitting on a sofa''. However, as argued by \cite{actionsarerelationchains},  such actions themselves require a more fundamental understanding of spatial relations,  for which decomposition into chains of spatial relations is important. Hence, relation understanding tasks should sit at the root of all other complex tasks. Many datasets~\cite{SceneGraphs, VSR} present a large number of spatial relations, with many overlapping meanings (such as \textit{( ``below'', ``beneath'')}, or \textit{(``adjacent to'', ``beside'')}. Both pairs are essentially describing the same preposition sense. Hence, the mixing of spatial relations with similar meanings results in redundancy.


Visual Relations can also be explicitly modeled as graphs~\cite{ashual2019specifying, ActionGenome, VisualGenome, VG50, SceneGraphs}, which can substitute the need for 3D information. This form of representation can also allow multiple spatial relations to co-exist, which may be useful in understanding complex scenes. While these works have shown strong performance in identifying low-level object relationships, understanding of higher-order relationships is still not evidently clear.

Recently, more advanced models have emerged \cite{chatterjee2024revision}, which create high-resolution datasets using text-to-image multimodal generation models, specially for spatial fidelity. 
\subsection{Video Datasets}
Many spatial relations have a dynamic nature, meaning they intrinsically involve motion (e.g. ``onto''), which cannot be represented by image datasets. Various works have proposed video datasets~\cite{iGibson2, ActionGenome, VidOR}, but they only cover the basic static positional prepositions (e.g. ``behind'', ``above'' and ``below''). \cite{VidOR} have a few additional static prepositions such as ``(facing) towards'', but overall, dynamic spatial and also temporal prepositions are severely under-researched. The CATER dataset~\cite{CATER} covers just the 3 most basic temporal prepositions (``before'', ``during'' and ``after'').

\subsection{Other related visual reasoning tasks}
Various other tasks are related to visual relationship reasoning, which require the use of both spatial and temporal cues to match visual features with labels for objects and relations. This includes tasks such as video grounding \cite{su2021stvgbert, li2022end, zeng2020dense} and visual question answering \cite{antol2015vqa, yusuf2022analysis}. Hence, many methods from visual relationship reasoning can be transferred to the above mentioned tasks and vice versa. 

%% file: content/3.dataset.tex
The STUPD dataset is a dataset for visual reasoning. It contains synthetically generated images and videos depicting spatial and temporal relations between objects. These relations are derived from the prepositions of the English language, which are words representing relations between different subjects within a sentence. The STUPD dataset provides 5,000 visual depictions of each preposition (in the form of either images or videos, depending on whether there is motion involved in the depiction of the relation), resulting in 150,000 samples corresponding to \textbf{spatial relations} (referred to as \textbf{Spatial-STUPD}). It also provides 50,000 videos corresponding to \textbf{temporal relations}(referred to as \textbf{Temporal-STUPD}). The data contains \textbf{realistic} interactions between objects of different kinds. The dataset is statistically balanced with respect to object distribution. The dataset can be used to pretrain models to perform visual reasoning better, as is demonstrated in this paper.  

\subsection{Predicate Vocabulary}
The Prepositions Project (TPP)~\cite{TPP}, a database of all prepositions in the English language, lists 373 prepositions in total. We use TPP as the source for our vocabulary, and select prepositions only from the two largest groups in TPP (spatial and temporal prepositions) for this paper. We first apply a structured filtering process on the list of all prepositions from TPP, the details of which are outlined in the appendix. Through the filtering process, we shortlisted \textbf{30 spatial prepositions and 10 temporal prepositions}. These prepositions act as predicate relations for our visual reasoning dataset. Spatial relation categories are divided into two subcategories -- static spatial relations (relations that do not involve relative movement between objects) and dynamic spatial relations (relations that involve movement of the objects). We describe all these relation categories, along with their definitions and context of usage, in Appendix \ref{tab:list of all preps}.  

\subsection{Setting and structure}
\subsubsection{Spatial dataset structure}
Consider a spatial relation triplet  \textit{$<$subject, predicate, object$>$}. For each predicate (relation) in the STUPD dataset, \textit{subject} and \textit{object} are represented by a collection of 3D objects. These 3D object templates (also referred to as prefabs) were selected from the ShapeNet dataset~\cite{chang2015shapenet}, which contains high-quality annotated 3D templates of real-life objects. The detailed curation process of the prefabs used is explained in Appendix \ref{overview of 3d prefabs used}. 






We group all object categories into 8 supercategories based on size and physical properties to simplify the types of interactions between different objects. These 8 supercategories are  \textit{small objects }(everyday objects that are small enough to be easily maneuvered), \textit{furniture, vehicles, person, large-scale grounded objects }(large heavy objects that are usually grounded, e.g. buildings)\textit{, containers, track }(roads and paths), and \textit{tunnels}. The idea behind supercategories is that categories within a supercategory have similar behavior of physical interaction. Overall, we curated 183 prefab instances varying across 45 object categories and 8 supercategories. An overview of the 3D prefabs used, along with other design choices are presented in Appendix \ref{overview of 3d prefabs used}. 

It should be noted that in STUPD, the representation of relation triplets \textit{(subject, predicate, object)} has a slightly different meaning than previous works. Certain predicate relations in our vocabulary such as \textit{(moving) up} and \textit{(scattered) all over (the scene)} describe the relation between the predicate and the subject category (and not any object category). Hence, the \textit{(object)} is empty for certain spatial relation categories. Note that the subject and/or the object can refer to multiple instances of the same physical object. 

\subsubsection{Temporal dataset structure.} Temporal predicates in  STUPD depict a relation between 2 events (a stretch of time where something occurs) or time points (a single moment of time). Consider the temporal relation triplet \textit{$<$Event/TimePoint A, relation, Event/TimePoint B$>$}. The challenging part of visual temporal relation representation is the visual depiction of events and time points. In this dataset, temporal relations are represented by means of videos, where events and time points are depicted using the spatial dataset generated. Each event is represented by a spatial relation that may occur over variable time spans. A static event simply means that there is an occurrence of a static relation a certain number of frames. On the other hand, time points are represented by single frame inside the temporal videos, and these are sampled from only static spatial events, since a single frame cannot represent the temporal nature of a dynamic spatial relation.  

\subsection{Dataset characteristics}

\subsubsection{Spatial-STUPD dataset characteristics}
All static spatial relations are generated as single RGB images(frames)(\(f=1\)), while dynamic spatial relations are generated as a collection of \(f=30\) consecutive RGB images (frames), which can be combined together to create a video depicting object interactions with dynamic movement. We synthetically generate 5,000 examples of each spatial relation using the Unity3D perception platform (\cite{borkman2021unity}), which allows the use of a physics engine to emulate realistic physical interactions between different objects.

\begin{figure*}
    \centering
    \includegraphics[width = \textwidth]{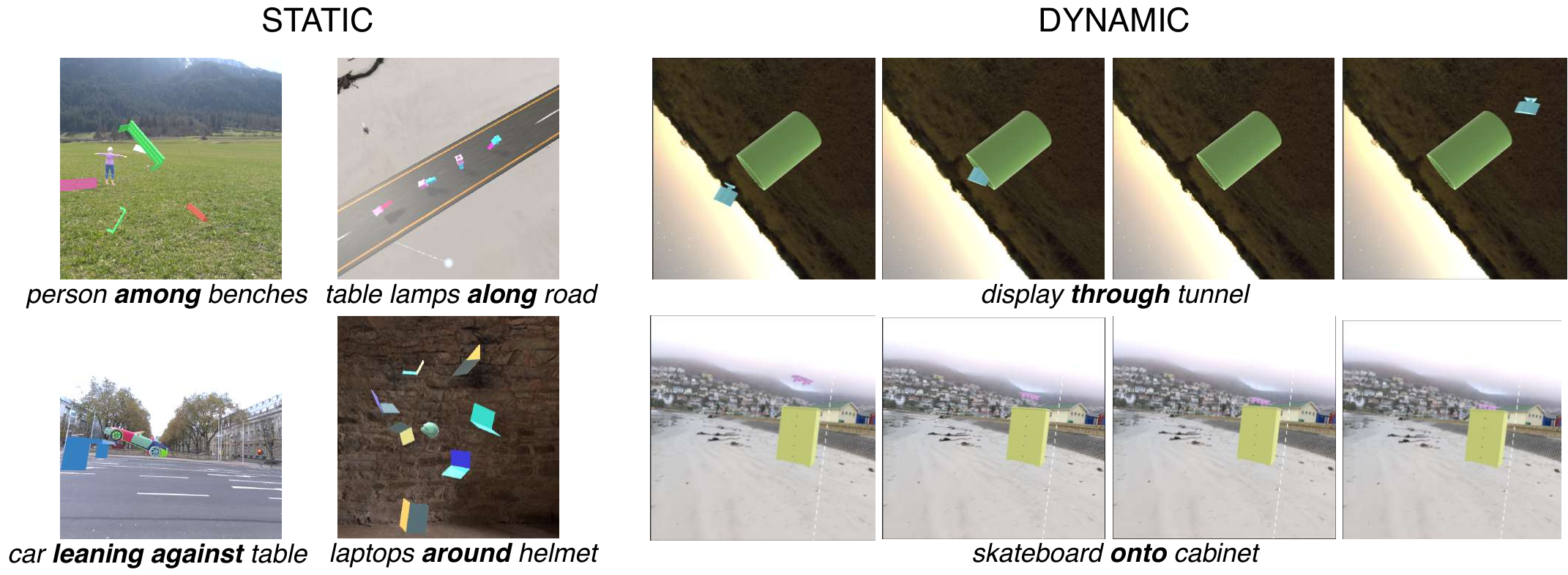}
    \caption{Some examples of Spatial-STUPD, which contains 30 spatial relations. These relations can be divided into two categories - static (involving no motion) and dynamic (involving relative motion between the subject and object)}
    \label{fig:STUPD visualization}
\end{figure*}

To ensure enough variance in the dataset, we randomize a variety of parameters of the generated images, such as the selection of the objects (in a constrained manner to only allow selective supercategory interactions, described above), object colors, the distance between the objects, the relative position and rotation of the objects, the perspective of the camera, and even the background of the image. All visual relations in the STUPD dataset are with respect to the camera's perspective, hence removing any ambiguity of perspective (such as left vs right). We provide annotations for each sample in the form of subject/object information (including category, supercategory, bounding box information and 3D coordinates), as well as the predicate category labels. All spatial relations are independent of each other. Hence each spatial interaction corresponds to only one predicate category.  Some examples of our dataset can be seen in Figure \ref{fig:STUPD visualization}, and Appendix figures \ref{fig:other static examples} and \ref{fig:other dynamic examples}. 

\subsubsection{Temporal-STUPD dataset characteristics} 
\begin{figure*}
    \centering
    \includegraphics[width=0.8\linewidth]{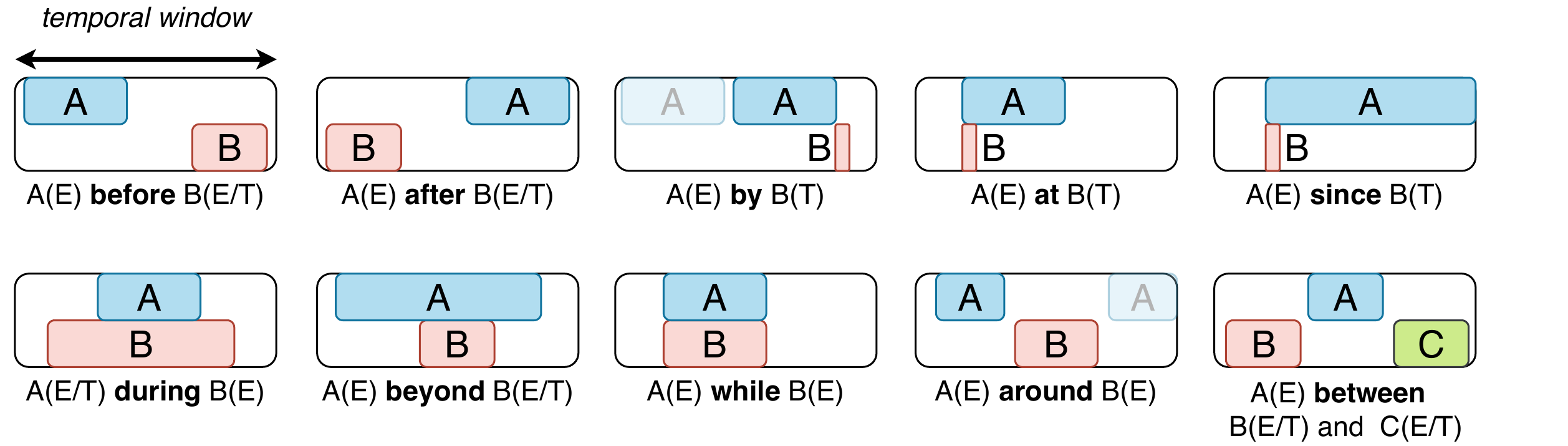}
    \caption{We propose 10 temporal relations representing interactions between different events or time points within a specified temporal window of \(W\) frames. Different temporal prepositions are used in specific contexts in English. For each relation, A, B, and/or C can be an event(E), time point(T) or either event or a time point(E/T). Each temporal relation can have multiple types of event/time point interactions. The translucent shade of certain events in the figure represents the possible variation in the point of occurrence.}
    \label{fig:temporal logic}
\end{figure*}

We generate pairs of videos of a constant length of \(W = 150\) frames (referred to as the temporal window), where each video corresponds to the occurrence of a single event or time point. An important characteristic of temporal relations is the overlapping nature of temporal relation predicates. Event/TimePoint interactions can represent multiple temporal relations simultaneously. For example, consider \textit{Event A} which occurs just after \textit{TimePoint B}. In this case, temporal triplets \textit{$<$Event A, after, TimePoint B $>$} and \textit{$<$Event A, around, TimePoint B$>$}  both apply.  Hence in the STUPD dataset, each temporal interaction may have multiple temporal relation categories associated.   An overview of all temporal relations is presented in Figure \ref{fig:temporal logic}.

\subsection{Statistics of the STUPD dataset}
\subsubsection{Spatial-STUPD.} This dataset's primary goal is to create a well balanced dataset, with a wide and balanced variety of \textit{subject-object} interactions. Firstly, each spatial and temporal relation has 5,000 samples each. As mentioned above, we constrain the interaction of supercategories to emulate real-world physical constraints and ensure similarity of size of \textit{subject} and \textit{object}. During dataset generation, we adjust the number of examples generated for each subject/object supercategory pair based on the total number of object categories interacting, so that individual category occurrences are more or less equal throughout the dataset. In Figure \ref{fig: stats}(a), we include the distribution of all supercategory occurrences in the STUPD dataset (including both subjects as well as objects). The frequencies are normalized by the number of prefab categories associated with each supercategory and presented as a fraction (percent). As can be seen, the normalized distribution of the majority of supercategories is more or less similar, with a couple exceptions, which arise out of conscious design choices, as described in Appendix \ref{overview of 3d prefabs used}.

\subsubsection{Temporal-STUPD.} 
\begin{figure}[]
\centering
\subfloat[]{%
  \includegraphics[clip,width=0.8\columnwidth]{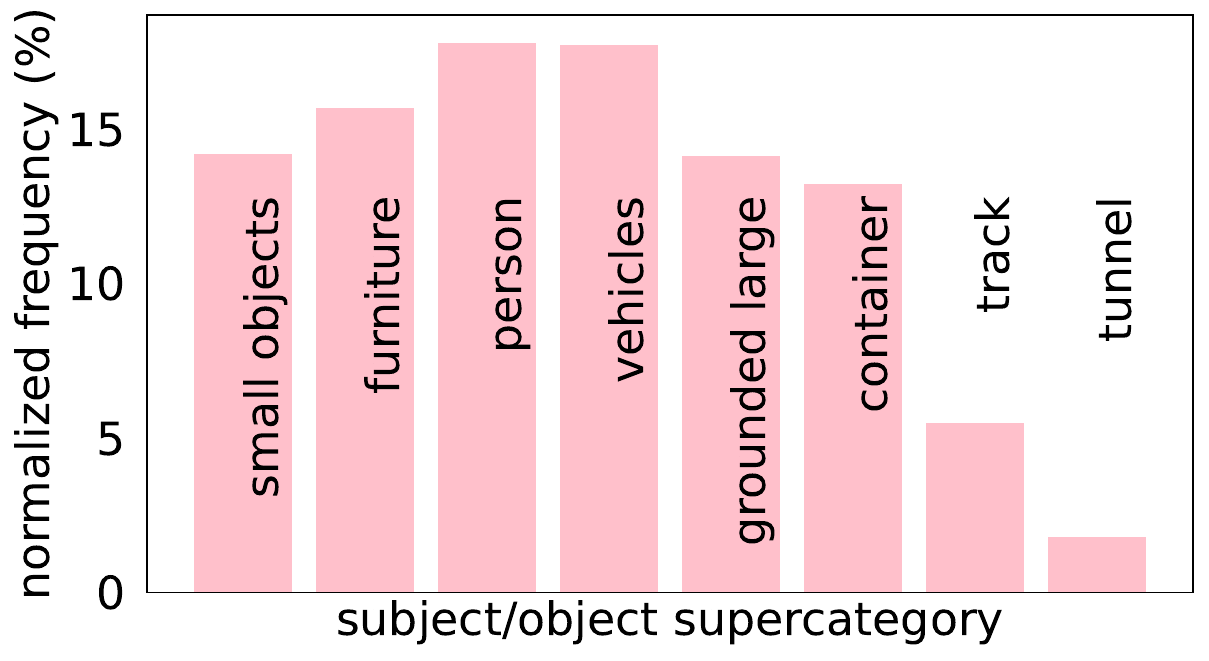}%
}

\subfloat[]{%
  \includegraphics[clip,width=0.8\columnwidth]{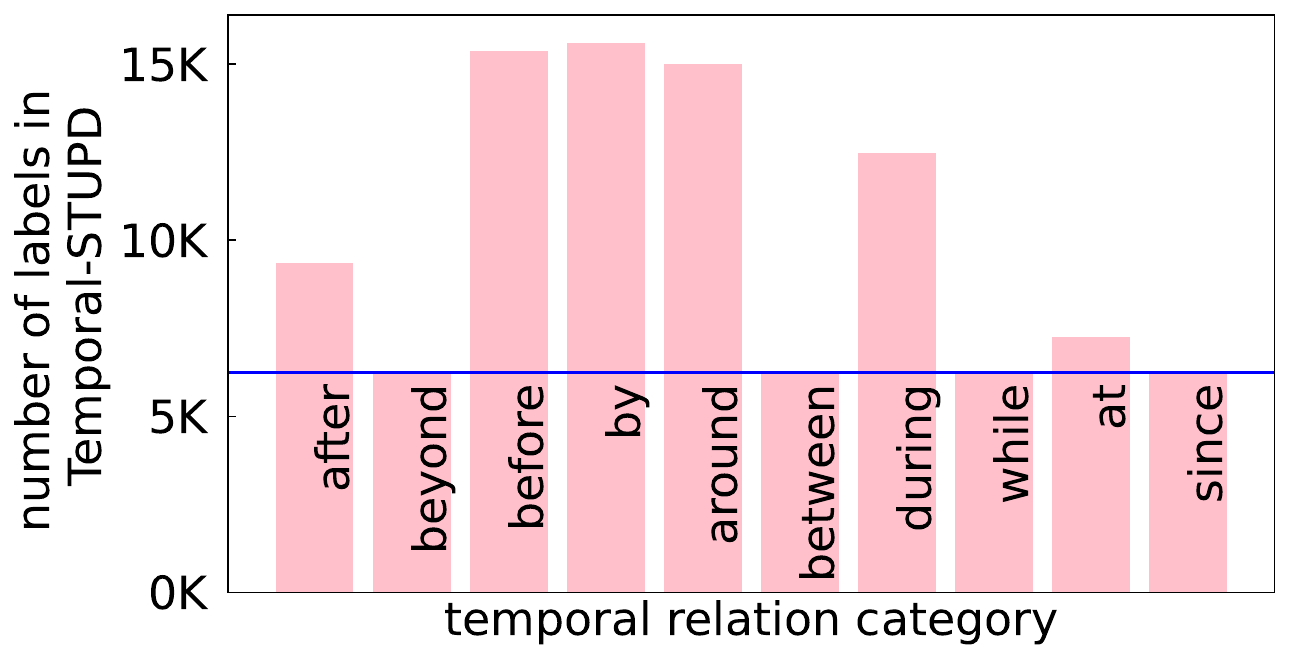}%
}

\caption{Dataset statistics. (a) The occurrence of prefab categories is roughly consistent throughout the dataset. (b) The blue line represents the minimum number of temporal relation occurrence. A single temporal interaction can have multiple temporal relation predicates associated. }
\label{fig: stats}
\end{figure}

Since Events/Time Points are randomly sampled from Spatial-STUPD, the distribution of Events/Time Points is similar to that in Figure \ref{fig: stats}(a). In Figure \ref{fig: stats}(b), we illustrate the occurrence of  different supercategories across the 50,000 data points. Each predicate has atleast 5,000 occurrences. However, because of the overlap between many temporal relations, many temporal predicates occur more frequently in the dataset. For instance, ``before" is a subset of ``by", and hence ``by" occurs whenever ``before" occurs, but not necessarily vice versa. Similary, ``while" is a subset of ``during" (related to two events occuring simultaneuously) and ``since" is a subset of ``at" (related to an event occurring at a particular time instance).

%% file: content/4.baselines.tex
\subsection{Spatial-STUPD baselines}
In this subsection, we aim to demonstrate that STUPD is an effective pretraining dataset for real world visual reasoning tasks. To demonstrate the effect of STUPD on real world visual reasoning tasks, we choose two real-world visual reasoning datasets - the SpatialSense Dataset \cite{Spatialsense} (to demonstrate performance on static spatial relations) and ImageNet-VidVRD \cite{shang2017video} (to demonstrate performance on dynamic spatial relations). 

\subsubsection{Selection of baseline models}
We choose various baselines to evaluate our dataset. These include two non-visual models (\textit{Language-based model}, which only takes verbal phrases as input, and \textit{Coordinate-based model}, which only takes the coordinates as input) and four vision-based classification models (\textit{DRNet}\cite{dai2017detecting}, \textit{VIPCNN} \cite{yikangvip}, \textit{PPRFCN} \cite{zhang2017ppr}, and \textit{VTransE} \cite{zhang2017visual}), along with two advanced vision-language models, CLIP \cite{clip} and VisualBERT \cite{li2019visualbert}. This collection of baseline models has been carefully selected to highlight the effect of different dataset components. For example, the language-based and coordinate-based models respectively highlight the statistical bias related to subject/object label distribution, and the effect of bounding box information, while isolating the effect of other components. It should be noted that we use already pre-trained versions of the CLIP and VisualBERT model. Also, since VisualBERT is a large model and thus tends to easily overfit on datasets of limited size, we use LoRA adapters \cite{hu2021lora} to fine-tune the model. 

 The models are evaluated on a single label predicate classification task (unlike previous approaches where a binary classification task is used to evaluate if a given relation triplet holds true). While this results in lower range performance values, it gives a more granular insight into the model's understanding of relations. The architecture of the models \textit{X} are adjusted according to the task and dataset used, and are thus referred to as`\textit{X-based}', to differentiate between the original architecture and the model used in this paper. Further details can be found in appendix \ref{training details}. 

\subsubsection{Preliminary Analysis of the dataset}
In Table \ref{STUPD_zero_shot}, we perform an overall as well as a preposition-wise analysis of the STUPD dataset, to get an understanding of the relative difficulty of individual categories. We use a pretrained CLIP \cite{clip} model to perform \textbf{zero-shot analysis} of the data, using natural language phrases as descriptions of relation triplets, accompanied by corresponding images. For dynamic prepositions, we uniformly sample 3 images, and aggregate the images to feed into the encoder. First off, we observe a slightly lower performance for dynamic prepositions than static prepositions, since the CLIP model has been pre-trained on images rather than videos.

We observe that certain relationships are easily interpreted by the pretrained CLIP model due to their clear visual depiction. For instance, the preposition Along (position), representing objects arranged along a track or path, is straightforward and unambiguous. In contrast, we also identify several low-performing relations, where a single image can depict multiple possible visual relationships, leading to ambiguity. Detailed insights into the preposition-wise performance are presented in Appendix \ref{zero shot analysis}

\subsubsection{Model performance on Spatial-STUPD}
Firstly, to validate the sanity of the STUPD dataset from a model training perspective, we train the baseline models on only the Spatial-STUPD dataset, and present the results in Table \ref{tab: only on stupd}. We note the suboptimal accuracy on the \textit{language-based} model. This is infact, a positive outcome. \textbf{Predicting the predicate based on only the subject and object category information represents imbalance and/or bias within the dataset.} A well-balanced dataset should produce low accuracy on this task, as is seen in the accuracy results. Next, we observe the best performance on Spatial-STUPD is achieved through the \textit{VisualBERT-based} model, followed by the \textit{coordinate-based} model, which is a relatively simple model. This demonstrates the higher importance of spatial coordinate/bounding box information over visual features. We also observe the higher performance on dynamic predicates in comparison to static predicates, with the exception of the \textit{DRNet-based} model. This indicates that dynamic data is loaded with more information for spatial reasoning, hence establishing the need for datasets with dynamic information. On the other hand, \textit{VisualBERT} \cite{li2019visualbert} model outperforms all models on static data, and overall on the entire STUPD dataset. This special suitability towards images rather than videos may be because of architectural design choices, but also because of the size of the model, leading to advanced reasoning.

\begin{table}[]
\centering
\resizebox{\linewidth}{!}{
\begin{tabular}{llll}
\hline
\textbf{Model} & \textbf{Overall} & \textbf{Static} & \textbf{Dynamic} \\ 
 & \textbf{Accuracy} & \textbf{Accuracy} & \textbf{Accuracy}\\\hline
Random & 3.34 & 3.34 & 3.34 \\ \hline
Language-based & 28.90                    & 26.76              & 31.66                    \\ 
Coordinate-based        & 75.60       & 72.54                  & 78.32           \\ 
VIPCNN-based & 64.24 & 61.52 & 70.37\\
PPRFCN-based  &68.19 & 66.41 & 69.47 \\
VTransE-based& 76.58  & 72.22 & 80.39 \\
DRNet-based          & 70.32                   & \textbf{81.35 }        & 60.70                   \\ 
CLIP-based & 72.86 & 79.29 & 67.23 \\
VisualBERT (LoRA) & \textbf{80.43} & 78.58 & \textbf{82.06} 

\end{tabular}
}
\caption{Visual reasoning performance trained on all 30 spatial relations in the Spatial-STUPD dataset. The values presented are accuracy metrics in percent. }
\label{tab: only on stupd}
\end{table}


\begin{table}[]
\centering
\resizebox{0.9\linewidth}{!}{%
\begin{tabular}{lll|lll}
\hline
\begin{tabular}[c]{@{}l@{}}\textbf{Static} \\ \textbf{}\end{tabular} & \begin{tabular}[c]{@{}l@{}}\textbf{Acc} \\ \textbf{}\end{tabular} & \begin{tabular}[c]{@{}l@{}}\textbf{Top-3} \\ \textbf{Acc}\end{tabular} & \begin{tabular}[c]{@{}l@{}}\textbf{Dynamic} \\ \textbf{}\end{tabular} & \begin{tabular}[c]{@{}l@{}}\textbf{Acc} \\ \textbf{}\end{tabular}  & \begin{tabular}[c]{@{}l@{}}\textbf{Top-3} \\ \textbf{Acc}\end{tabular} \\ \hline

Random & 3.3 & N/A & Random & 3.3 & N/A \\ 
\textbf{Overall} & \textbf{23.3} & \textbf{37.8} &\textbf{ Overall }& \textbf{20.31 }& \textbf{33.18} \\ \hline
Above & 28.6 & 49.1 & Around & 10.3 & 24.0 \\
On & 8.4 & 16.7 & Onto & 7.7 & 26.5 \\
\begin{tabular}[c]{@{}l@{}}\textbf{Along}\\ \textbf{(position)}\end{tabular} & \textbf{34.8} &\textbf{65.8}& \begin{tabular}[c]{@{}l@{}}Out \\ (of)\end{tabular} & 4.2 & 13.6 \\
All over & 10.3 & 22.2 &  \textcolor{red}{Up} &  \textcolor{red}{0.0} &  \textcolor{red}{0.0} \\
\textbf{Among} & \textbf{54.6 }& \textbf{68.1 }& Along & 52.2 & 71.1 \\
\begin{tabular}[c]{@{}l@{}} \textcolor{red}{Against}\\  \textcolor{red}{(leaning)}\end{tabular} &  \textcolor{red}{7.2} &  \textcolor{red}{16.5} & Into & 5.6 & 17.7 \\
Outside & 38.3 & 53.4 &\textbf{Towards}& \textbf{47.4} & \textbf{72.8} \\
In front of & 8.1 & 17.0 &  \textcolor{red}{Through} &  \textcolor{red}{2.2} &  \textcolor{red}{7.8} \\
 \textcolor{red}{Behind} &  \textcolor{red}{4.3} &  \textcolor{red}{12.2} & From & 3.5 & 13.5 \\
Between & 12.0 & 21.7 &  \textcolor{red}{With} &  \textcolor{red}{2.0} &  \textcolor{red}{9.6} \\
Inside & 30.3 & 52.2 & Against & 14.3 & 36.8 \\
Below & 39.6 & 56.4 & \textbf{By} & \textbf{85.8} &\textbf{90.1} \\
\textbf{Around} & \textbf{50.1} & \textbf{70.8} &\textbf{Off}& \textbf{60.6} &\textbf{ 71.4} \\
 \textcolor{red}{Beside} &  \textcolor{red}{4.2} &  \textcolor{red}{12.6} & \begin{tabular}[c]{@{}l@{}}Into\\ (crash)\end{tabular} & 21.0 & 35.2 \\
 &  &  & Down & 6.8 & 16.4 \\
 &  &  & Over & 7.5 & 31.2
\end{tabular}%
}
\caption{Zero-Shot \textbf{preposition-wise} and overall performance using a pretrained \textbf{CLIP} \cite{clip} \textbf{model}. Values are accuracy percent values. We report both accuracy (top-1) and top-3 accuracy. The three best-performing and worst-performning prepositions in each category are highlighted in bold and red, respectively.}

\label{STUPD_zero_shot}
\end{table}

\subsubsection{Generalization to real-world settings}

We propose STUPD primarily as an effective pretraining dataset before transfering on real-world dataset. To demonstrate the effect of pretraining a model on STUPD (followed by fine-tuning on real world datasets, except in the case of zero-shot CLIP evaluation) , we compare the results of pretraining on various datasets. We consider two real world datasets (\textit{SpatialSense} \cite{Spatialsense}, to demonstrate the results on static relations in the form of  images), and  \textit{ImageNet-VidVRD}\cite{shang2017video}, for dynamic relations in the form of videos). For each dataset, we compare pre-training on different suitable datasets, such as the \textit{ImageNet} dataset \cite{deng2009imagenet}(for the \textit{SpatialSense} dataset)/\textit{KINETICS-400} dataset \cite{carreira2017quo}(for the \textit{ImageNet-VidVRD} dataset), and the \textit{CLEVR} dataset \cite{CLEVR} (which is similar to Spatial-STUPD in relation settings). 


It can be seen that Spatial-STUPD dataset, when used as a pretraining dataset for visual relationship reasoning tasks, improves performance on real-world datasets. On the other hand, CLEVR does not lead to a significant increase in performance in comparison to from-scratch training in most cases. Finally, it can be seen that ImageNet pretraining infact does not help improve performance in any significant manner. This is also observed in the case of pretraining on the Kinetics-400 dataset. Overall, STUPD is well aligned for various visual relation reasoning tasks, in comparison to other similar synthetic datasets, as well as very large general pretraining datasets like ImageNet. An exception can be seen in the case of CLIP pretraining, where direct fine-tuning of the CLIP model on SpatialSense leads to the best results. This is, however, a result of overfitting, since the CLIP model used was already pretrained on other datasets.

It is evident from the results that effective visual relationship reasoning results not only from rich visual cues (from ImageNet or Kinetics400 pretraining), but also from additional cues such as relative spatial positioning, bounding box, and language-based information. It can also be noticed that the jump in accuracy after pretraining is much more pronounced in the case of ImageNet-VidVRD training than SpatialSense training. This indicates the importance of dynamic information for effective visual reasoning.

\begin{table}[]
\centering
\resizebox{\linewidth}{!}{
\begin{tabular}{lllll}
\multicolumn{2}{l}{(SpatialSense Training)} & \multicolumn{3}{l}{\textbf{Pretraining dataset}}  \\ \hline
 \textbf{Model}      & \textbf{no pre-} & \textbf{Image-} & \textbf{CLEVR} & \textbf{Spatial} \\ 
 
    & \textbf{training} & \textbf{Net} &  & \textbf{STUPD} \\ \hline
Random               & 16.67                   & 16.67             & 16.67          & 16.67          \\ 
{CLIP (0-shot)} & 17.80 & 14.92 & 28.56 & \textbf{44.26} \\ \hline
Language-based       & \textbf{43.13}                   & N/A                 & 43.04          & 42.91          \\
Coordinate-based     & 47.45                   & N/A                & 47.62          & \textbf{49.59} \\
VipCNN-based         & 41.17                   & 41.94             & 41.11          & \textbf{44.28} \\
PPRFCN-based         & 44.12                   & 42.61             & 42.08          & \textbf{44.98} \\
VTransE-based        & 49.81                   & 49.85             & 46.98          & \textbf{50.84} \\
DRNet-based          & 51.93                   & 52.54             & 52.84          & \textbf{54.28} \\
CLIP-based & \textbf{54.77} & 44.42   & 48.90 & 52.16 \\ 
VisualBERT (LoRA) & 56.88 & 48.50 & 52.16 & \textbf{62.12}\\
\end{tabular}}
\caption{Effect of Spatial-STUPD pretraining on the SpatialSense \cite{Spatialsense} dataset. Fine-tuning the model on the real world dataset after pretraining on STUPD leads to the best results.} The values presented are accuracy metrics in percent.
\label{tab: pretraining on spatialsense} 

\end{table}

\begin{table}[]
\centering
\resizebox{\linewidth}{!}{
\begin{tabular}{lllll}
\multicolumn{2}{l}{(ImageNet-VidVRD training)}  & \multicolumn{3}{l}{\textbf{Pretraining dataset}}                               \\ \hline 
\textbf{Model}  & \textbf{no pre-} & \textbf{KINETI-} & \textbf{CLEVR} & \textbf{Spatial} \\ 
   
 & \textbf{training} & \textbf{CS-400} & & \textbf{STUPD} \\ \hline
    
Random               & 10.00                   & 10.00                 & 10.00          & 10.00          \\ \hline
Language-based       & 54.35                   & N/A                   & \textbf{55.25} & 54.71          \\
Coordinate-based     & 54.49                   & N/A                   & 52.11          & \textbf{54.79   }       \\
VipCNN-based         & 50.68                   & 50.54                 & 58.44          & \textbf{86.95} \\
PPRFCN-based         & 51.72                   & 51.87                 & 49.87          & \textbf{62.64} \\
VTransE-based        & 56.60                   & 56.88                 & 64.64          & \textbf{73.97} \\
DRNet-based          & 57.98                   & 57.29                 & 68.07          & \textbf{87.29} \\
CLIP-based & 61.41  & 55.12 & 66.47 & \textbf{71.26} \\
VisualBERT (LoRA) & 71.12 & 63.08 & 58.62 & \textbf{76.30}

\end{tabular}}
\caption{Effect of Spatial-STUPD pretraining on the ImageNet-VidVRD \cite{shang2017video} dataset. Again, fine-tuning the model on the real world dataset after pretraining on STUPD leads to the best results.}  The values presented are accuracy metrics in percent.
\label{tab: pretraining on vidvrd}
\end{table}

\subsection{Temporal-STUPD baselines}

 We formulate a simplified task for Temporal-STUPD to demonstrate that models benefit on real-world datasets when pretrained on this dataset. An obvious domain which can benefit directly from pretraining on Temporal-STUPD is visual question answering (VQA) \cite{manmadhan2020visual}, because of the intersection of visually grounded spatial informatin and diverse language-based event descriptions. For the purpose of finetuning a Temporal-STUPD finetuned model on a real world VQA dataset, we choose the \textit{NeXT-QA} dataset \cite{xiao2021next}, which contains structured language-based questions accompanying videos of everyday events. 
We slightly modify the traditional VQA grounding task to match the structure of NeXT-QA with Temporal-STUPD. \textit{NeXT-QA} presents language annotations in the form of multiple choice questions. Corresponding to temporal prepositions, NeXT-QA presents three types of visual questions -- An event occuring before another event, an event occuring after another event, and an event occuring at a specific moment in the video. To match the triplet format Temporal-STUPD (\textit{$<$Event/TimePoint A, relation, Event/TimePoint B$>$}), we split the question-answer pair string by the relational word -- giving us a viable proxy to Events/TimePoints A and B. This temporal relation triplet extraction method is described through examples in the Appendix \ref{pretraining on nextqa details}.

In the modified task, given Event A and Event B, a model performs a binary classification to predict whether the correct relation is `before' or `after', due to certain limitations of the NeXT-QA dataset (details can be found in Appendix \ref{pretraining on nextqa details}). We choose 3 baseline models to evaluate the effect of pretraining on Temporal-STUPD before finetuning on NeXT-QA -- a language-only model, two other vision-language models presented in \cite{xiao2021next}, EVQA \cite{antol2015vqa} and STVQA \cite{jang2019video}, along with a modified CLIP-based \cite{clip} architecture. We slightly modify the architectures as originally proposed in \cite{xiao2021next} to adapt the models to the new task involving predicate classification. More details can be found in Appendix \ref{baseline model design process: appendix}. Although the task is heavily simplified, the effect of the Temporal-STUPD dataset is clearly evident.

\subsubsection{Model performance on Temporal-STUPD}

Table \ref{tab: pretraining on nextqa} presents the results of pretraining experiments on Temporal-STUPD.  Similar to the Spatial-STUPD, we observe sub-optimal performance in the language-based model, and only marginal improvement as a result of pretraining. This is expected, since a language-only model only highlights the statistical distribution of the labels. However, when language and spatio-temporal fatures are combined, we observe significant improvements in the performance of the models when pretrained on Temporal-STUPD, in comparison to when models are trained on NeXT-QA from scratch. On the other hand, we observe that models, in-fact, perform worse  than a model trained from scratch, when pretrained on KINETICS-400. This reinforces two key observations -- rich real-world spatial information is not sufficient (or helpful) in effective temporal relation reasoning, and that pretraining on a diverse set of rich temporal relation dataset (like Temporal-STUPD) boosts real-world training significantly.

\begin{table}[]
\centering
\resizebox{0.9\linewidth}{!}{
\begin{tabular}{llll}
\multicolumn{2}{l}{(NeXT-QA training)} & \multicolumn{2}{l}{\textbf{Pretraining Dataset}}                \\
\multicolumn{2}{l}{(Temporal)} &   \\
\hline
\textbf{Model}              & \textbf{no pre-} & \textbf{KINETICS-} & \textbf{Temporal} \\ 

  & \textbf{training} & \textbf{400} & \textbf{STUPD} \\

\hline
Language-based      &         50.75       &        N/A              &     \textbf{51.37 }          \\
EVQA-based              &          62.69       &       57.52                &         \textbf{71.42}       \\
STVQA-based   &      64.24`      &        58.76        &             \textbf{70.52}  \\
CLIP-based & 48.27 & 61.02 & \textbf{76.34}
\end{tabular}}

\caption{Effect of Temporal-STUPD pretraining on the NeXT-QA \cite{xiao2021next} dataset. The values presented are balanced accuracy metrics, which represent the average of class-wise recall values, in percent. This metric is chosen because of the unbalanced nature of the two relevant classes in NeXT-QA. For this reason, we also omit the random baseline, since in skewed distributions, random predictions lose their value.}
\label{tab: pretraining on nextqa}
\end{table}


%% file: content/5.limitations.tex

The STUPD dataset was designed with simplicity in mind. A preposition can have multiple senses, sometimes with subtle differences in meaning or usage in different contexts. In the case of spatial relations, we restrict context of usage by limiting subjects and objects to physical objects, thus allowing us to group different senses into a single preposition. Further works may focus on creating visual datasets to disambiguate the subtle differences of meanings of a preposition. Another design choice was to limit the types of objects to at most 2 types (categories) per image, for simplicity. However, this somewhat limits with number of potential prepositions included, as some comparative prepositions require 3 types of objects in order to be depicted properly. An example is \textit{as far as}, which depicts a comparison between two distances. This cannot be represented through interaction between only two objects.

Finally, while 3D information (such as relative depths of objects) is readily available in STUPD due to its synthetic nature, this was not fully utilized in this paper, primarily in order to compare the results with previous works fairly. Future works may examine whether and how 3D information may help with certain reasoning tasks.

%% file: content/6.conclusion.tex

Static representations such as image based datasets are not sufficient for machine learning systems to fully understand spatial relations well. Spatial relations have many subtle characteristics such as relative movement, velocity, direction, orientation, which can only be fully justified through flexible dynamic representations such as synthetic based videos. In this paper, we introduced a novel dataset which aims to cover the subtle differences between different spatial relations through simple object interactions. Through various experiments, it is evident that the dynamic nature of senses helps model identify relations better. Our studies also demonstrate the nature of spatio-temporal learning in 3D deep learning models. It is observed that models initially rely more on spatial cues, but slowly learn about temporal cues as well, and the combination of spatio-temporal cues results in higher accuracy. 

Although this dataset consists of simple object interactions, we hope that it can be used to make models understand more complex scene structures, such as nuanced contexts of preposition use in the English language, or for understanding the underlying dynamics of actions better in various action recognition tasks. 

%% file: content/Appendix.tex
\subsection{Glossary of various terms used in this paper}

Many terms have been used in this paper, which may be related, and hence confusing to the reader. In Table \ref{tab: glossary}, we clearly define a few key words related to our work. 

\begin{table*}[]
\centering
\begin{tabular}{ll}
\textbf{Term}                          & \textbf{Definition}                                                                                                                                           \\ \hline \hline  
\textit{\textbf{relation}}             & any interaction between two physical objects or time points.                                                                                                                                              \\
\textit{\textbf{predicate}}             & another term for \textit{relation}, in the context of a relation triplet.  \\

\textit{\textbf{preposition}}          & words that depict different kinds of relations.                                                                                                                                                           \\
\textit{\textbf{prepositional sense}}  & \begin{tabular}[t]{@{}l@{}}a particular meaning of a prepositional word, especially in a specific context. \\ the same prepositional word can have different meanings in different contexts.\end{tabular} \\
\textit{\textbf{spatial preposition}}  & a preposition describing the relation between two physical objects.                                                                                                                                       \\
\textit{\textbf{static preposition}}   & \begin{tabular}[t]{@{}l@{}} a spatial preposition that involves no motion, i.e. the object(s) involved are  \\ stationary   \end{tabular} \\

\textit{\textbf{dynamic preposition}}  & \begin{tabular}[t]{@{}l@{}} a spatial preposition that involves motion, i.e. the object(s) involved are \\ non-stationary   \end{tabular}                                                                                                             \\
\textit{\textbf{temporal preposition}} & a preposition describing the relation between two time points or time periods.                           \\
\hline
\end{tabular}
\caption{Glossary for the various terms used in this paper.}
\label{tab: glossary}
\end{table*}

\subsection{Predicate Vocabulary}
\label{prepositions filtering process}
We use The Prepositions Project (TPP)~\cite{TPP}, as a starting point for selecting the prepositions for the purpose of visual reasoning representation in STUPD. 
The process we implemented to shortlist our list of spatial and temporal prepositions is outlined here.
\begin{enumerate}
    \item All spatial and temporal prepositions were ranked by the frequency of occurrence in standard text corpora, which acts as a proxy to an indicator of word usage in the English Language. 
    \item We set a threshold to the frequency of occurrence, and discard all prepositions with a frequency smaller than the threshold since these prepositions are either archaic or have a more frequently used synonym.
    \item Then we discard certain prepositional senses based on rules as follows.
    \begin{enumerate}
        \item The preposition should not represent a negative sense (e.g. the absence of an object), since they cannot be depicted visually. 
        \item  The preposition should not represent interaction between three or more object categories, since this represents a collection of multiple spatial relations, including pairwise relation between objects. This would increase the complexity of our dataset.
        \item The preposition should not be context-specific (i.e. related to usage with respect to only selected types of objects), since these prepositions create bias within the dataset.
    \end{enumerate}
    \item One goal of the STUPD dataset was to represent generalizable visual relations, that do not depend on any metric scales of distance (such as kilometers, meters, etc.) and time (seconds, minutes, etc.). For example, the temporal sense ``\textit{under}'' is used to represent an event that lasts for less than a specific amount of time (e.g. The noodles were ready in \textit{under} two minutes). Hence we discard prepositional senses that require comparison with specific metrics.

    \item Many of the remaining prepositions are similar in meaning and can be grouped together (e.g. \textit{(``on'', ``on top of'', ``upon'') or (``below'', ``beneath'', ``under'', ``underneath'')}. Hence, we group all prepositions into categories having mutually exclusive definitions,
\end{enumerate} 

During the filtering process, we make one specific exemption, where we chose to discard two spatial relation categories - "\textit{to the left of}" and "\textit{to the right of}". This is because of the perspective problem presented by these prepositions - the viewer's (camera's) perspective of these relations can be the opposite of the \textit{subject}'s perspective, and also different for the \textit{object}'s perspective. This creates ambiguity in the dataset, which is harmful to visual reasoning detection tasks.

This filtering process led to a finalised shortlist of 40 distinct prepositional senses, consisting of 30 spatial prepositional senses and 10 temporal prepositional senses. The 30 spatial prepositional senses are derived from  26 different spatial prepositions, and consist of 14 static spatial relations and 16 dynamic spatial relations. All 40 prepositional senses have distinct meanings and contexts of usage. The list of all prepositions proposed in STUPD, along with their specific definitions/contexts are presented in Table \ref{tab:list of all preps}.


\begin{table*}[]
\centering
\begin{tabular}{lll}
\textbf{Prepositional Type} & \textbf{Preposition} & \textbf{Definition} \\
\hline \hline
Spatial - static & \textit{\textbf{above}} & at a higher level than \\
 & \textit{\textbf{against}} & \begin{tabular}[t]{@{}l@{}}in physical contact with (something), so as to be \\ supported by\end{tabular} \\
 & \textit{\textbf{all over}} & everywhere \\
\textit{} & \textit{\textbf{along}} & extending in a more or less horizontal line on \\
 & \textit{\textbf{among}} & \begin{tabular}[t]{@{}l@{}}situated more or less centrally in relation to (several other \\ things)\end{tabular} \\
 & \textit{\textbf{around}} &  on every side of, so as to encircle (someone or something) \\
 & \textit{\textbf{behind}} & \begin{tabular}[t]{@{}l@{}}to the far side of (something), typically so as to be hidden \\ by it\end{tabular} \\
 & \textit{\textbf{below}} & at a lower level than \\
 & \textit{\textbf{beside}} & at the side of (next to) \\
 & \textit{\textbf{between}} & at, in, or across the space separating (two objects) \\
 & \textit{\textbf{in front of}} & \begin{tabular}[t]{@{}l@{}}in a position just ahead or at the front part of someone or \\ something else \end{tabular}  \\
 & \textit{\textbf{inside}} & situated with the boundaries of (a container) \\
 & \textit{\textbf{on}} & physically in contact with and supported by (a surface) \\
 & \textit{\textbf{outside}} & situated beyond the boundaries or confines of (a container) \\
 \hline
Spatial - dynamic & \textit{\textbf{against}} & moving in the opposite direction of \\
 & \textit{\textbf{along}} & moving in a constant direction on \\
 & \textit{\textbf{around}} & \begin{tabular}[t]{@{}l@{}}so as to pass (a place or object) in a curved or approximately \\ circular  route.\end{tabular} \\
 & \textit{\textbf{by}} & so as to go past \\
 & \textit{\textbf{down}} & movement from a higher to a lower point \\
 & \textit{\textbf{from}} & \begin{tabular}[t]{@{}l@{}}indicating the point in space at which a motion starts in the \\ direction away\end{tabular}   \\
 & \textit{\textbf{into}} & \begin{tabular}[t]{@{}l@{}}expressing movement with the result that something becomes \\ surrounded   by (a container)\end{tabular} \\
 & \textit{\textbf{into}} & \begin{tabular}[t]{@{}l@{}}expressing movement (collision) with the result that \\ something  makes   physical contact with something else\end{tabular} \\
 & \textit{\textbf{off}} & moving away from \\
 & \textit{\textbf{onto}} & moving to a location on the surface of \\
 & \textit{\textbf{out of}} & movement from within the boundaries of (a container) \\
 & \textit{\textbf{over}} & expressing passage or trajectory directly upwards from \\
 & \textit{\textbf{through}} & moving in one side and out of the other side of \\
 & \textit{\textbf{towards}} & moving in the direction of \\
 & \textit{\textbf{up}} & movement from a lower to a higher point \\
 & \textit{\textbf{with}} & movement in the same direction as \\
 \hline
Temporal & \textit{\textbf{after}} & in the time following (an event or another period of time) \\
 & \textit{\textbf{around}} & occurring slightly before or after a given point of time \\
 & \textit{\textbf{at}} & expressing the exact time when an event takes place \\
 & \textit{\textbf{before}} & in advance of the time when \\
 & \textit{\textbf{beyond}} & happening or continuing after (a specified time or event) \\
 & \textit{\textbf{between}} & in the period separating (two events or points in time) \\
 & \textit{\textbf{by}} & indicating a deadline or the end of a particular time period \\
 & \textit{\textbf{during}} & at a particular point in the course of (an event) \\
 & \textit{\textbf{while}} & coinciding at the same time of; meanwhile \\
 & \textit{\textbf{since}} & \begin{tabular}[t]{@{}l@{}}in the intervening period between (the time mentioned) and \\ the time under consideration, typically the present.\end{tabular}
\end{tabular}
\caption{List of all prepositional senses used in STUPD and their definitions. In total, we derive 40 prepositional senses from 34 distinct prepositions, of which there are 30 spatial prepositional senses (derived from 26 distinct prepositions) and 10 temporal prepositional senses.}
\label{tab:list of all preps}
\end{table*}

\subsection{Overview of 3D prefabs selection}
\label{overview of 3d prefabs used}

The primary goal of STUPD is to create an effective pre-training dataset for real-world visual reasoning tasks. As such, we first filter out frequently occurring objects in three real-world visual reasoning datasets (the Visual Genome dataset, \cite{VisualGenome}, the SpatialSenses dataset~\cite{Spatialsense} and the ImageNet VidVRD dataset~\cite{shang2017video} and categorize all object categories into clusters based on similarity into supercategories (e.g., ``car'', ``motorcycle'' and ``bicycle'' can be categorized into \textit{vehicles}). 

Based on these clusters, we select related object prefabs from the ShapeNet dataset ~\cite{chang2015shapenet}, including but not limited to the categories identified from the real-world visual reasoning datasets mentioned above (continuing the aforementioned example, we select other \textit{vehicle} prefabs as well, such as ``bus'', ``boat'' and ``train''). This is done to increase statistical variation in the STUPD dataset. Some other 3D prefabs, such as \textit{person}, \textit{tunnel}, and \textit{track (road)} were added to the STUPD repository from various community platforms that provide free 3D prefabs for open use. All prefabs are manually resized to maintain relative size (e.g. a soda can is smaller in size than a table lamp). The ranges of relative size for objects are chosen based on the average size of the object in the real world. 

\begin{figure*}[]
\centering
\includegraphics[width=.45\textwidth]{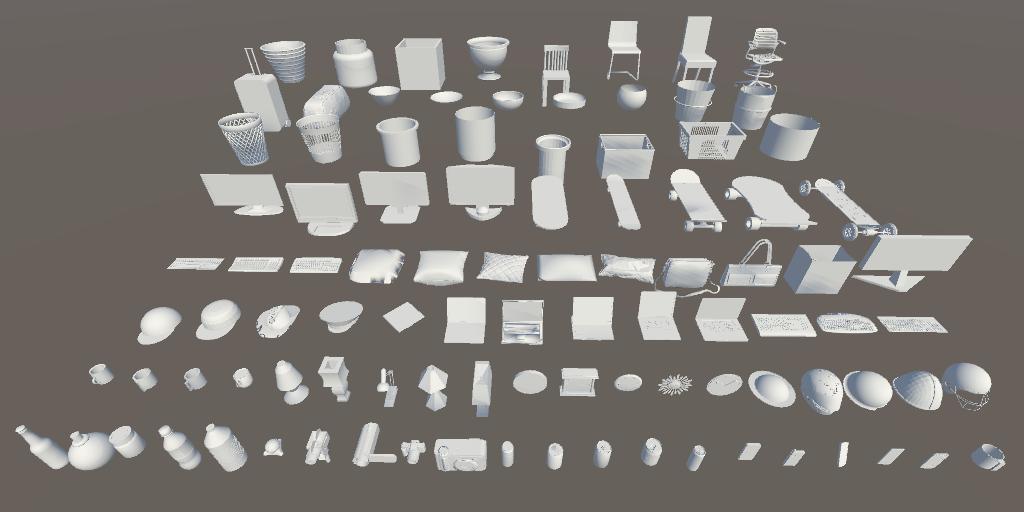}
\includegraphics[width=.45\textwidth]{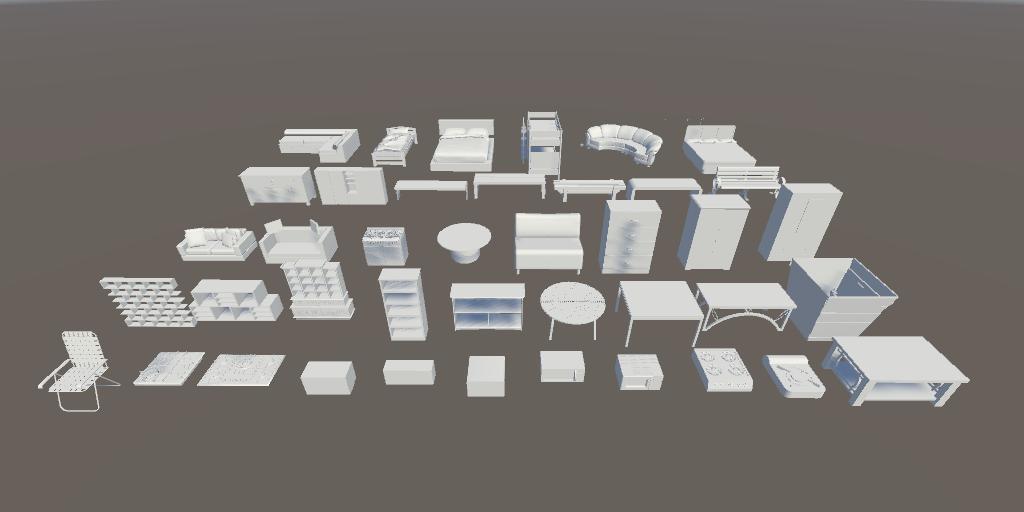}

\includegraphics[width=.45\textwidth]{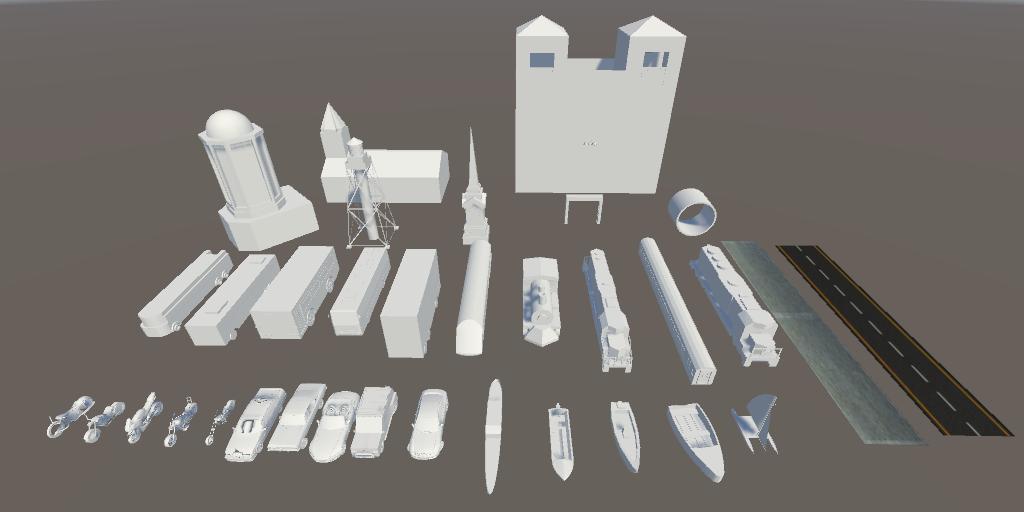}
\includegraphics[width=.45\textwidth]{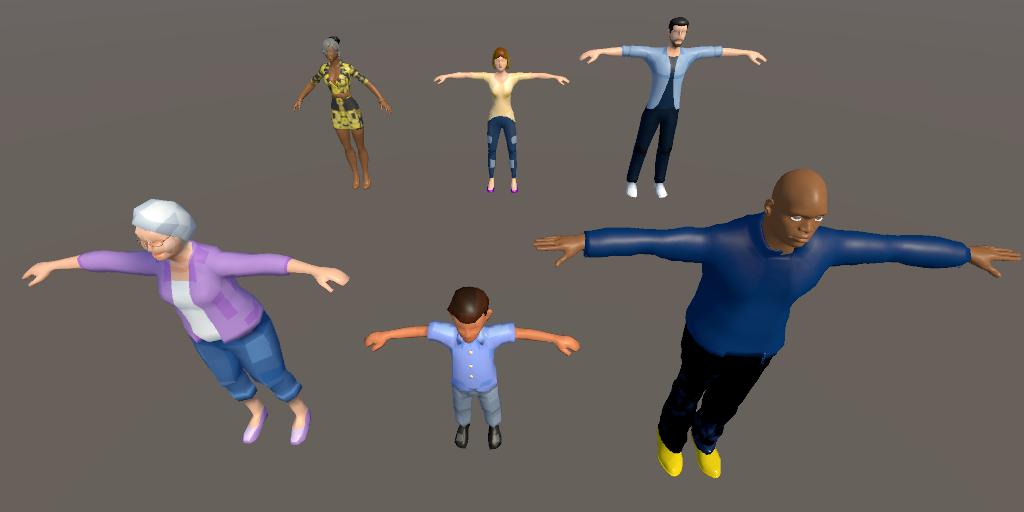}

\caption{Overview of 3D prefabs used. We curate a total of 183 prefabs across 45 categories and 8 supercategories. We also try to have a balanced set of \textit{person} prefabs to address certain ethical concerns (bottom right). (Also see Figure \ref{fig:ethnic distro}) }
\label{fig:prefabs overview}
\end{figure*}

\begin{figure}
    \centering
    \resizebox{\linewidth}{!}{
    \includegraphics{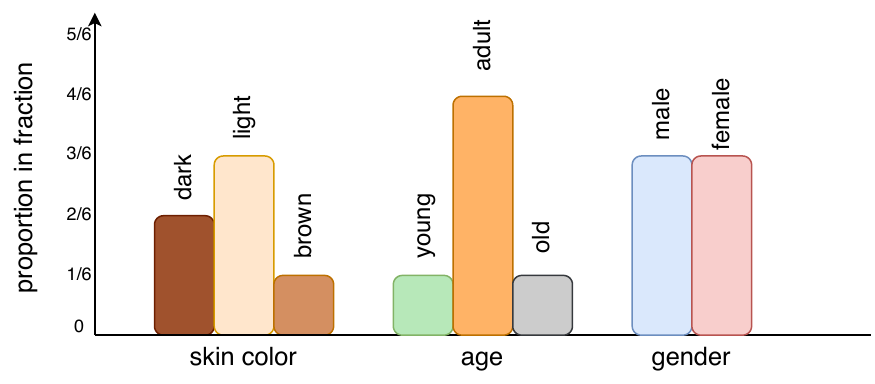}}
    \caption{Ethnic distribution statistics for the \textit{person} prefab. }
    \label{fig:ethnic distro}
\end{figure}

We also try to address certain ethical concerns in our object collection. Specifically, to avoid bias in visual relation detection tasks, we curate 6 instances of \textit{person}, with an equal number of male and female characters, with varying skin color, age, height, and clothing. Figure~\ref{fig:prefabs overview} depicts all the prefabs that were used for the generation of the STUPD dataset. Some statistics of the race (skin color), age and gender are depicted in Figure \ref{fig:ethnic distro}.




        

We allow only selective pairs of supercategory interactions for each spatial relation, to maximize realism in the dataset. For example, consider the relation predicate \textit{$<$above$>$}. It is unlikely to come across an image example of \textit{$<$building above person$>$}, since a building cannot hover in the air.  Additionally, we also filter out supercategory pairs with a large difference of size. For example, we avoid interaction between a train and a coffee mug, since the size difference is so large, that both cannot simultaneously fit into the same image without compromising the visibility of at least one of these objects. Thus our supercategory pair  filtering process involves filtering out all infeasible supercategory \textit{$<$subject, object$>$} combinations based on physical constraints and differences of size. 

Further, to induce realism in the STUPD dataset, we curated 10 background images containing a variety of sceneries, such as blue sky, ocean, grass fields, desert, cityscape, beach, indoor settings, etc. The backgrounds are carefully chosen so there are no visual distractions, such as objects that might confuse computer vision models. Additionally, we include variation in lighting direction and intensity, thus allowing us to create a realistic shadowing effect as well. The images are not 2D images, but 3D spheres overlaid around the object interaction space during synthetic generation. Hence, with a combination of rotating camera viewpoints and varying lighting effects, a large variety of background images are constructed, ensuring a diverse background set. Moreover, the selection of backgrounds is completely randomized, to ensure that no statistical bias is introduced while training models on STUPD.

Our approach gives the STUPD dataset various advantages over previous synthetically generated visual reasoning datasets, such as CLEVR~\cite{CLEVR} and Rel3D~\cite{Rel3d}. Since we use real-world objects and backgrounds in our synthetic dataset, along with realistic physical interactions, rather than simple objects such as cubes or spheres, our dataset can transfer knowledge into real-world visual reasoning tasks much more efficiently. Additionally, while we filter out unrealistic object interactions, we simulate interaction between many object categories that are rarely found in real-world datasets such as Visual Genome~\cite{VisualGenome}, and SpatialSenses~\cite{Spatialsense} (for example \textit{$<$car on car$>$}).  The diverse range of object interactions in STUPD will allow visual reasoning models to learn spatial relations by learning only the interaction between objects, and not the type of objects (which is a kind of bias inherent in many real-world visual reasoning datasets). There are many potential benefits of using the STUPD dataset in real-world settings. A simple example is $<$\textit{person (moving) down$>$}. Because not many examples of this event may be found  in real-world visual reasoning tasks, it can cause visual reasoning models to fail to detect an  event such as a $<$\textit{person falling down from a building}$>$ in real-world settings. Our dataset aims to fill this knowledge gap for visual reasoning models.

While we try to balance the occurrences of different objects in STUPD, there are some anomalies (as seen in Figure \ref{fig: stats} (for example, the slightly lower occurrence of grounded large objects and larger containers, as well as the low occurrence of tracks and tunnels. These arise because of conscious design choices. Some key observations that led to such design choices are presented below. 

\begin{enumerate}
    \item Supercategories `track' and `tunnel' have lower frequencies because of their association with only a small number of spatial relations (such as ``(moving) \textbf{along} track" and ``subject (passing) \textit{\textbf{through}} a tunnel". 
    \item It can be seen that the frequency of `small objects' is slightly lower than others. This is a conscious design choice, because of the size mismatch between other supercategories, having much larger sizes (such as buildings, vehicles, or furniture). We, however, try to maintain balance by including appropriate numbers of interactions between the larger objects within the small objects supercategory and other supercategories. 
\end{enumerate}

\subsection{Spatial-STUPD dataset}

\begin{figure*}
    \centering
    \includegraphics[width=0.8\textwidth]{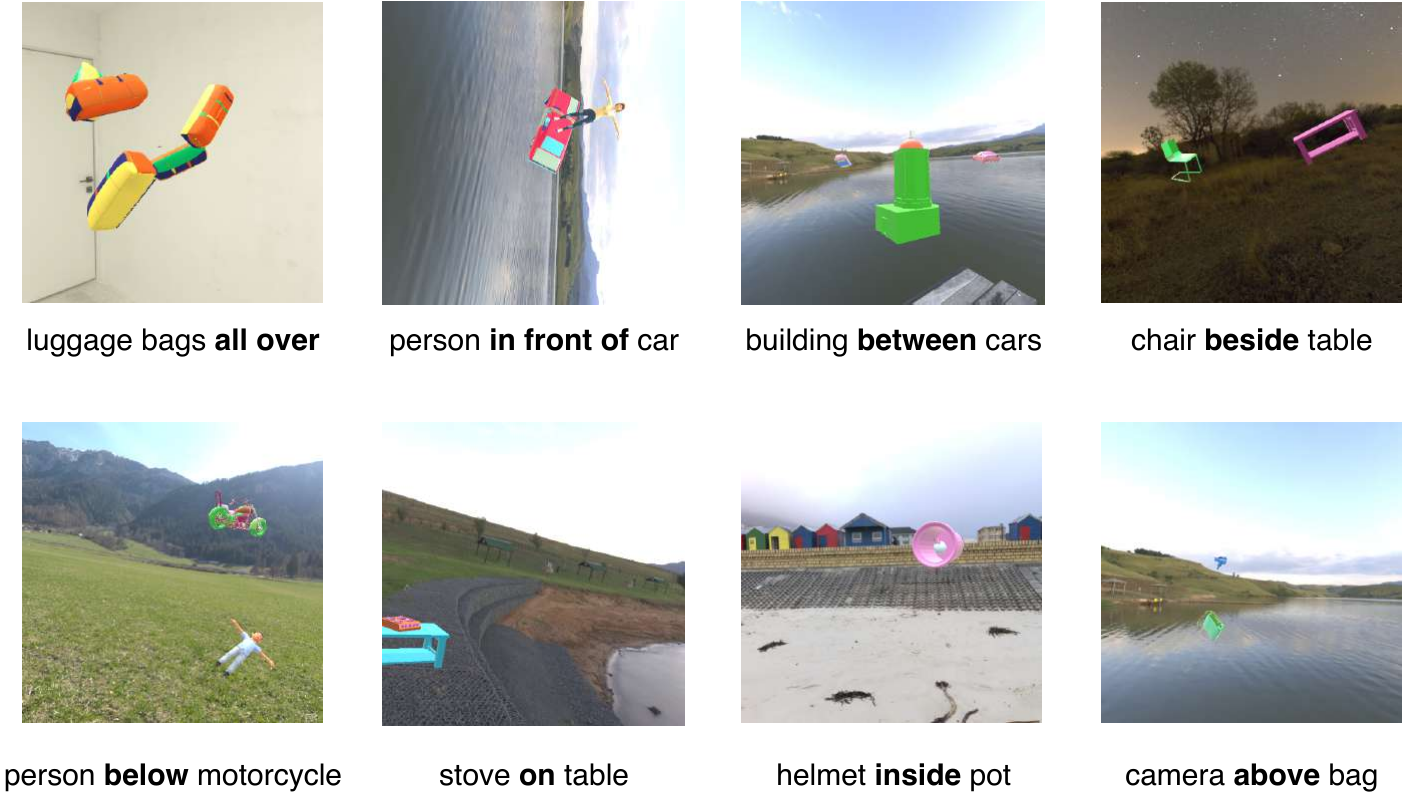}
    \caption{Some other static spatial relations in STUPD}
    \label{fig:other static examples}
\end{figure*}

\begin{figure*}[htp]
    \centering
    \includegraphics[width=0.8\textwidth]{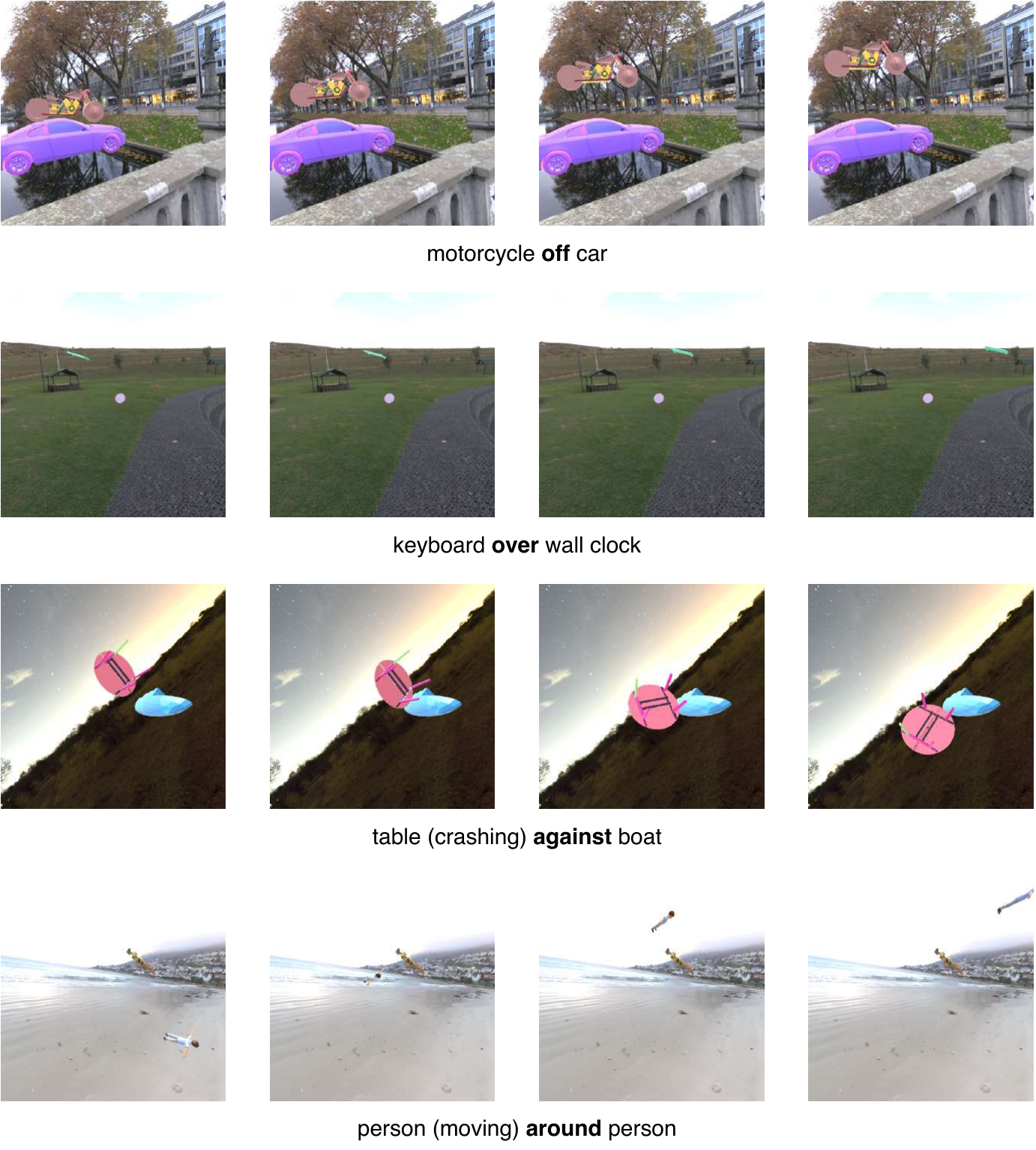}
    \caption{Some other dynamic spatial relations in STUPD. Note that some predicates may have multiple senses (definitions) in different context. We represent all 30 spatial senses with a unique label in the STUPD dataset.}
    \label{fig:other dynamic examples}
\end{figure*}

We present some other examples of spatial relations in this section. Refer to Figure \ref{fig:other static examples} for examples of static spatial relations and Figure \ref{fig:other dynamic examples} for examples of dynamic spatial relations. 

\subsection{Temporal-STUPD dataset}

\label{temporal stupd dataset generation}
In the Temporal-STUPD dataset, the events vary from \(f' = 15\) to \(f'= 45\) frames, and are represented uniformly. 
In the video, only the frames corresponding to the occurrence of the event/time point are visually illuminated, and the remaining sections of the video are blacked out. This approach gives certain advantages to us. 
\begin{enumerate}
    \item Since all temporal relations are independent of the nature of events themselves (for example, in \textit{$<$Event A, after, Event B$>$}, Event A and B can be any spatial relation occurrence), our method segregates the events and time points into different video streams, allowing temporal reasoning models to effectively understand this independence. 
    \item Our approach allows us to scale the temporal interaction to $>$2 events/time points. For example, consider the temporal predicate \textit{$<$between$>$}, which involves interaction between 3 events/time points (\textit{$<$Event A occurs between Event B and Event C$>$}). Hence we provide annotations of each temporal relation in the form of information about the temporal relation category, and information about \textit{Event A}, \textit{Event B}, and\textit{ Event C} (where \textit{Event C} field can be empty), such as which spatial event it corresponds to, along with the frame numbers corresponding to the start and end of any event/time period. 
\end{enumerate}

An event of \(f'\) frames (\(f' \epsilon [15,45]\)) can be represented by a static event (where the same image is spread across \(f'\) frames, representing the occurrence of a static event for a particular length of time) or a dynamic event (where the original \(f=30\) frames are interpolated by uniform sampling (if \(f'$<$=f\)) or spherical interpolation (if \(f'$>$f\)). Spherical interpolation ensures that the corresponding video is smooth even if the frame rate for a dynamic spatial interaction is reduced.

\begin{figure*}
    \centering
    \includegraphics[width=\textwidth]{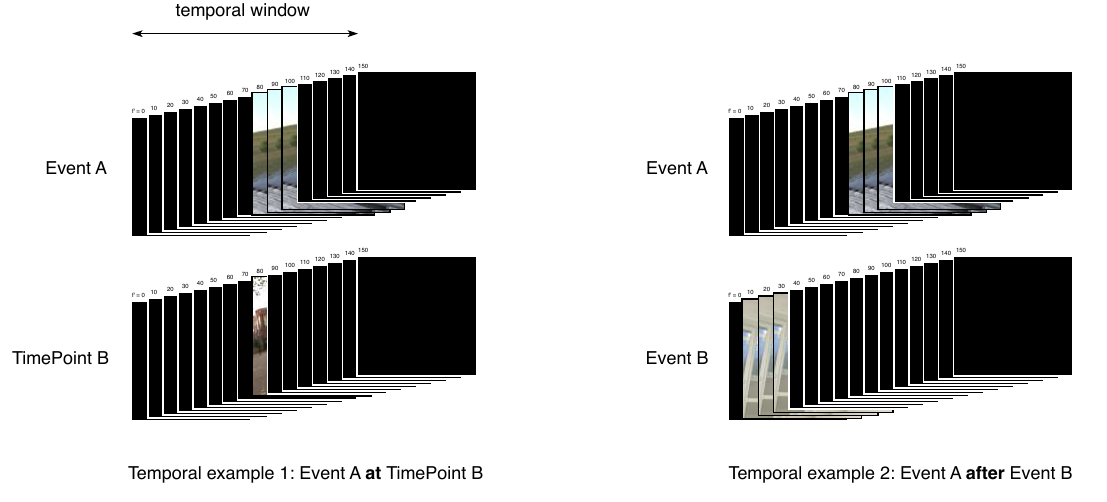}
    \caption{Examples of temporal relations in STUPD. Each temporal relation can be a relation in time between 2 or more events or time points. Each event (spanning over multiple frames in the temporal window) is represented through either a static or dynamic spatial relation, while each time point (a single frame in the temporal window) is represented by a static spatial relation.}
    \label{fig:temporal examples}
\end{figure*}

An overview of temporal relations in STUPD is presented in Figure \ref{fig:temporal examples}.

\subsubsection{Pretraining experiment design}
\label{pretraining on nextqa details}

We use the NeXT-QA dataset as an example of a real-world dataset, and perform experiments to see if pretraining on Temporal-STUPD improves real-world temporal relation reasoning. However, most standard VQA datasets have a question-answer pair format. In order to pretrain models on Temporal-STUPD, followed by finetuning on NeXT-QA, we introduce a temporal relation triplet extraction heuristic, that utilizes the linguistic structure of sentences involving the words `before' and `after', which may be described as \textit{$<$Question asking about event A$>$ before/after $<$description of event B$>$? $<$Description of event A$>$}. Using this structure, we convert the natural-language question-pair into structured relation triplets as proposed in this paper. 

For example, consider the question-answer pair from NeXT-QA: \textit{``What happens \textbf{after} the lady pushes the girl? The girl slides down.''} A simple sentence manipulation logic based around the word ``after" yields the temporal relation triplet \textit{$<$the girl slides down, after, the lady pushes the girl$>$}. 
In the case of NeXT-QA, we notice that while question/answers involving \textit{A preceding B}, and  \textit{B preceding A} involve just two prepositional relations (before and after), the examples involving \textit{A occuring at a specific moment} containa a diverse range of natural language structures, not necessarily involving any temporal relations. Hence, for structural integrity, we only evaluate pretraining effects of Temporal-STUPD for the questions involving \textit{A before B}, and \textit{A after B}. An analysis of the triplet extraction heuristic reveals that most of language-based context is preserved in the triplet across all temporal examples in the NeXT-QA dataset.

We perform classification between only `before' and `after' in NeXT-QA. To match the video format in NeXT-QA, we combine the video pairs in Temporal-STUPD by overlaying the videos on each other, resulting in a video with a sequence of distinct events. 

\subsection{Choice of real-world dataset}
To demonstrate that Spatial-STUPD is an effective pretraining dataset for spatial reasoning tasks, we evaluate fine-tuning of various models pretrained on STUPD, on ImageNet-VidVRD (for dynamic spatial relations) and SpatialSense (for static spatial relations). While other datasets, especially for static spatial relations, could also be included in our analysis. Unfortunately, at the time of writing this paper, Visual Genome~\cite{VisualGenome} and VRD~\cite{VRD} datasets were not accessible due to server issues. Therefore, we were unable to retrieve the necessary data and incorporate them into our research. We acknowledge that these datasets could have potentially provided valuable insights and comparisons for our study. However, the unavailability of these datasets prevented us from including them in our analysis.

In the case of Temporal-STUPD, we evaluate fine-tuning of various model on the NeXT-QA dataset \cite{xiao2021next}. This dataset has certain advantages over other VQA dataset. It covers certain simple temporal relations (before/after/during), unlike most VQA datasets that focus on spatial relations only, or do not provide sophisticated annotations for the temporal relations involved. Secondly, the structured annotation format makes it easy to convert natural-language question-answer pairs into triplets, that directly match the format of annotations provided in STUPD.

\subsection{Dataset reproducibility}
Throughout the dataset generation process, many parameters were assumed. To provide greater flexibility to the users of this dataset, we release all dataset generation scripts, environments, prefabs, parameter value sheets, and documentation to allow users to recreate or adjust values (and/or logic of the spatial/temporal dataset generation). Thus, while we release 150,000 spatial relations and 50,000 temporal relations, in theory, a lot more images can be generated as seen fit by users. We observed that generating the dataset in small batches with different seed values in UNITY results in a diverse dataset. 


\subsection{Zero-Shot analysis on STUPD}
\label{zero shot analysis}

In Table \ref{STUPD_zero_shot}, we perform a preposition wise analysis using zero-shot classification through a pretrained CLIP encoder. In each category of prepositions (dynamic and static), we observe the highest-performing prepositions as well as the lowest performing prepositions. 

The highest performing prepositions majorly constitute of unambiguous visual representations, which cannot be confused with another preposition. The prepositions \textit{Along (position)}, \textit{Around} and \textit{Among} fall in this category. Similarly, some of the lowest performing prepositions arise because of visual ambiguity (e.g. \textit{(moving) up} can be confused with \textit{(moving) down}, or other static relations such as \textit{besides}. A similar case can be made for \textit{Behind} (vs \textit{In front of}).

However, other extremities are likely a result of a skewed representation of relations in real world datasets. For example, Visual Genome \cite{VisualGenome}, which is used for pretraining the CLIP model does contain relations related to \textit{Towards} (through many compound relations), but does not represent other relations like \textit{Through} or \textit{Off}. The lowest performing relations are mostly from relations that are not well represented in current visual relation datasets. 

\subsection{Training details}
\label{training details}

\subsubsection{Training splits for the STUPD dataset}

We intend this dataset to be randomly split into training, validation and/or testing splits, and hence do not provide any explicit labels for this. This design choice arises due to many reasons. First, since this dataset aims towards being a pretraining dataset, performance metrics on STUPD should not be strictly optimized, and only used as a rubric to understand a models understanding of spatial/temporal relations. Secondly, a random split is representative of the balanced distribution of STUPD. Due to this reason, we also do not report the accuracy of predicate classification on the STUPD dataset during pretraining stages. 

\subsubsection{Baseline model design process}
\label{baseline model design process: appendix}
\paragraph{Spatial-STUPD baselines} 
In this paper, we design the task to be a single-label predicate classification task. Various visual relation detection models (including the models used in this paper for baselining) showcase model performance through a binary classification task, where the information of the subject and object, as well as the predicate is fed as input to the models, and the task is to evaluate if the relationship triplet holds true. However, a binary classification approach does not clearly indicate the performance of models, because the answer being only one of two options. In this paper, our approach is to only feed models (except for CLIP and VisualBERT) with information about subject and objects, and predict the predicate as a single-label classification task. While the accuracy achieved through this technique is much lower than a binary classification task, the granularity allows us to understand the differences between different models and tasks clearly. To adjust for the modified task, the architectures of the various models has been adjusted such that no information of the predicate is fed as input, and the final layer is replaced with a single-label classifier.

In the case of CLIP \cite{clip} and VisualBERT \cite{li2019visualbert}, a slightly different approach is used. Since CLIP-based models are pretrained to match images with natural-language based tags, we try to maintain the naturality of the relation triplet semantics. Hence, we first expand the relation triplet into a more linguistically coherent phrase, before tokenizing the phrase and feeding it into the model. For example, consider an example 

\textit{$<$ ball towards chair$>$}

is converted into

\textit{``A video of a ball moving towards the table."}

The image is fed into the model as an uncropped image. It should be noted that we use a pretrained backbone for the CLIP model (based on OpenAI's VIT B-32 model, pretrained on the MS-COCO, Visual Genome and the YFCC100M datasets \cite{clip})

It should also be noted that the models used for baselining in this paper are primarily designed for image data. Hence, in order to support training over dynamic data, such as Imagenet-VIDVRD \cite{shang2017video}, we modify the architectures by replacing 2D convolutional backbones with ResNet-based 3D convolutional layers \cite{tran2018closer}, optionally pretrained over the KINETICS-400 dataset \cite{carreira2017quo} (as opposed to the ImageNet dataset in the traditional 2D convolutional network). 

For pretraining, we select only a part of the STUPD dataset. For all pretraining tasks, we select predicate categories that are shared by both the real-world dataset and the STUPD dataset. This leads us with a 6-way single-label classification task for the SpatialSense dataset, and a 10-way single label classification task for the ImageNet-VidVRD dataset. A similar strategy is followed for the CLEVR dataset. In the case of ImageNet dataset, we only pretrain the convolutional backbone of the model. For a fair comparison, we train all models with the same hyperparameters and number of epochs of training. All pretraining tasks (except ImageNet pretraining) are carried out for 2 cycles, while finetuning on the real-world task is carried out for 5 cycles. The ImageNet pretraining is carried out for 20 cycles.

\paragraph{Temporal-STUPD baselines}
The corresponding predicate classification task for Temporal-STUPD involves a binary classification problem (corresponding to the classes \textit{`before'} and \textit{`after'}), because of certain limitations of the concerned real world dataset (NeXT-QA) (see \ref{pretraining on nextqa details}). We select three models to demonstrate the effect of pretraining on Temporal-STUPD, namely the language-based model (similar to Spatial-STUPD), the EVQA-based model \cite{antol2015vqa}, and STVQA-based model \cite{jang2019video}. The architectures of EVQA and STVQA models have been slightly modified to adapt to the relation predicate classification task. The language component, which is fed as input to all these models involves parsing both Events/TimePoints A and B (in the temporal relation triplet) through a word embedding model, and concatenated into one single vector. Then this vector is parsed through an LSTM model \cite{yu2019review}. In the case of EVQA and STVQA models, the input video features (stack of images) is parsed through a Resnet18-3D  model, and the fully connected layer activations are dot-multiplied with language embeddings, before being fed into the final fully-connected classification layer. Except the KINEMATIC-400 pretrained model, all other models take 10 frames from sampled uniformly from the video (ie, 10 temporally equidistant frames). For the backbone pretrained on KINEMATIC-400, we only use 3 frames, because KINEMATIC-400 pretrained models sued 3 input channels only.

\subsubsection{Effect of Training Data size on performance}
To demonstrate that the size of the STUPD dataset is sufficient for models to learn spatial relation information, we sample the dataset in different ratios (starting from 5\% of a 135K samples in SpatialSTUPD for training, all the way to 100\%). These sample sets are chosen randomly. At each sample, we train a DRNet-based model and evaluate on a 15K sample set randomly separated for testing purpose). The results of the classification task are shown below.

\begin{table}[]
\resizebox{\linewidth}{!}{
\begin{tabular}{c|cccccc}
Dataset Sample Ratio (\%)& 5 & 12.5 & 25 & 50 & 75 & 100 \\ \hline
Accuracy (\%) & 6.0 & 12.5 & 33.1 & 46.9 & 68.8 & \textbf{70.3}
\end{tabular}
}
    \caption{Accuracy on Spatial-STUPD on different training dataset sizes, when trained on the DRNet-based model}
    \label{tab:dset size comparison}
\end{table}

The accuracy achieved on the complete (100\%) dataset matches with the results presented in Table \ref{tab: only on stupd}. We observe a gradual saturation towards the peak performance as we increase the dataset size. Hence it can be confirmed that the dataset size (150K for spatial-STUPD) is sufficient for a visual relation dataset. Moreover, this dataset is the largest among all other dataset in its domain, making it the best candidate for pretraining visual reasoning models using synthetic dataset.

\subsection{datasheets for dataset}

In this section, we provide the datasheets for the STUPD dataset. 

\subsubsection{Motivation}
The questions in this subsection are primarily intended to encourage dataset creators to clearly articulate their reasons for creating the dataset and to promote transparency about funding interests.

\begin{enumerate}
    \item \textbf{For what purpose was the dataset created?} 
The STUPD dataset is a large-scale synthetic dataset primarily designed to act as an efficient pretraining dataset for visual reasoning tasks in real-world settings. It also is a balanced dataset representing a wide variety of spatial and temporal relations, all with distinct definitions. 


\end{enumerate}

\subsubsection{Composition}
The questions in this section are intended to provide dataset consumers with the information they need to make informed decisions about using the dataset for their chosen tasks. Some of the questions are designed to elicit information about compliance with the EU’s General Data Protection Regulation (GDPR) or comparable regulations in other jurisdictions.

\begin{enumerate}
    \item \textbf{What do the instances that comprise the dataset represent (e.g.,documents, photos, people, countries)?} RGB images of standard dimensions 512x512, depicting interaction between different types of 3D objects. 
    \item \textbf{How many instances are there in total (of each type, if appropriate)?} There are 200K instances in total - 150K instances for spatial relations and 50K instances for spatial temporal relations. The spatial relation instances are evenly distributed across 30 prepositional relations, each consisting of 5,000 examples each. Out of the 30 relations, there are 14 static relations, which are represented by 1 image per instance, and 16 dynamic relations, represented by 30 images per instance (depicting a video of 30 frames total). Each temporal instance contains one or more streams of videos containing 150 images each. We provide all videos in the form of images. In total STUPD is a collection of 17,920,000 images. 
    \item \textbf{Does the dataset contain all possible instances or is it a sample (not necessarily random) of instances from a larger set?} No, the dataset has been curated (generated) from scratch. 
    \item \textbf{What data does each instance consist of?} Each data instance is in the form of an RGB image.
    \item \textbf{Is there a label or target associated with each instance?} We provide annotations for each data instance, providing information about the objects in the image(s), their nature (which category and supercategory they belong to), the type of spatial/temporal relation that the image is part of, the 2D bounding boxes of all objects in the image, and the 3D spatial coordinates of the center of the objects in the image.
    \item \textbf{Is any information missing from individual instances?} No, the images have been provided as generated by Unity through scripts, which we publish for open-source access. 
    \item \textbf{Are relationships between individual instances made explicit?} The annotations make clear which images belong to which data instance. 
    \item \textbf{Are there recommended data splits (e.g., training, development/validation, testing)?} We recommend a random split for training purposes. The goal of STUPD dataset is to act as an effective pretraining dataset. 
    \item \textbf{Are there any errors, sources of noise, or redundancies in the dataset?} There are a few images with minor errors. For example, some objects may be partially or, in extreme cases, completely out of the field of view. Another source of error may be the overlapping of 3D objects, which may look unnatural if compared with real-world settings. However, in most cases, the core relation represented by the image still holds true, and a combination of information from images and annotations contains all the required information for effective visual relation reasoning. 
    \item  \textbf{Is the dataset self-contained, or does it link to or otherwise rely on external resources (e.g., websites, tweets, other datasets)?} Our dataset is self-contained. In fact, we also provide all required scripts and environments for users to regenerate or modify our dataset as they see fit for their own purpose. 
    \item \textbf{Does the dataset contain data that might be considered confidential (e.g., data that is protected by legal privilege or by doctor–patient confidentiality, data that includes the content of individuals’ non-public communications)?} No.
    \item \textbf{Does the dataset contain data that, if viewed directly, might be offensive, insulting, threatening, or might otherwise cause anxiety?} No. 
\end{enumerate}

\subsubsection{Collection Process}
In addition to the goals outlined in the previous section, the questions in this section are designed to elicit information that may help researchers and practitioners to create alternative datasets with similar characteristics. We only address relevant questions in this section.
\begin{enumerate}
    \item \textbf{How was the data associated with each instance acquired? } The data was generated by Unity3D through a custom scripting process.
\end{enumerate}

\subsubsection{Preprocessing/cleaning/labeling}
The questions in this section are intended to provide dataset consumers with the information they need to determine whether the “raw” data has been processed in ways that are compatible with their chosen tasks.

\begin{enumerate}
    \item \textbf{Was any preprocessing/cleaning/labeling of the data done (e.g., discretization or bucketing, tokenization, part-of-speech tagging, SIFT feature extraction, removal of instances, processing of missing values)?} We removed redundant information from the meta files generated by Unity during the dataset generation process, and only retained the information necessary for visual reasoning tasks. 
    \item \textbf{Was the “raw” data saved in addition to the preprocessed/cleaned/labeled data (e.g., to support unanticipated future uses)?} Yes, users can request for the raw metadata files generated by Unity by contacting the authors. Additionally, we provide data value sheets that contain all information about parameters that users can use to regenerate the dataset as originally intended. 
    \item \textbf{Is the software that was used to preprocess/clean/label the data available?} Yes, Unity3D is an openly available software. We also release the environment files that will aid users to quickly setup their application to generate data. 
\end{enumerate}

\subsubsection{Uses}
The questions in this section are intended to encourage dataset creators to reflect on the tasks for which the dataset should and should not be used. By explicitly highlighting these tasks, dataset creators can help dataset consumers to make informed decisions, thereby avoiding potential risks or harms.
\begin{enumerate}
    \item \textbf{Has the dataset been used for any tasks already?} No, we introduce this dataset for the first time in this paper. 
    \item \textbf{Is there a repository that links to any or all papers or systems that use the dataset?} Yes, will be provided later. 
    \item \textbf{What (other) tasks could the dataset be used for?} Potentially, this dataset can be used to transfer knowledge from the synthetic domain to the real-world domain, and to learn the dynamics of physics from the interaction of different kinds of objects. 
    \item \textbf{Is there anything about the composition of the dataset or the way it was collected and preprocessed/cleaned/labeled that might impact future uses?} No. 
\end{enumerate}

\subsubsection{Distribution}
Dataset creators should provide answers to these questions prior to distributing the dataset either internally within the entity on behalf of which the dataset was created or externally to third parties.

\begin{enumerate}
    \item \textbf{Will the dataset be distributed to third parties outside of the entity (e.g., company, institution, organization) on behalf of which the dataset was created?} No. 
    \item \textbf{How will the dataset will be distributed (e.g., tarball on website, API, GitHub)?} We will provide a link where users can download the entire dataset from a server. Links will be provided later. 
    \item \textbf{Will the dataset be distributed under a copyright or other intellectual property (IP) license, and/or under applicable terms of use (ToU)?} The dataset will be provided under the \textbf{CC BY-NC-SA 4.0} license. The Unity scripts and environment files will be released under the  \textbf{GNU General Public License 3.0}.
    \item \textbf{Do any export controls or other regulatory restrictions apply to the dataset or to individual instances?} No.

\end{enumerate}

\subsubsection{Maintenance}
The questions in this subsection are intended to encourage dataset creators to plan for dataset maintenance and communicate this plan to dataset consumers.

\begin{enumerate}
    \item \textbf{Who will be supporting/hosting/maintaining the dataset?} The authors will support and maintain the dataset. Access to the dataset via servers will be supported by A*STAR internal funding. 
    \item \textbf{How can the owner/curator/manager of the dataset be contacted (e.g., email address)?} Users can contact the authors with their official email addresses. 
    \item \textbf{Is there an erratum?} (to be updated later)
    \item \textbf{Will the dataset be updated (e.g., to correct labeling errors, add new instances, delete instances)?} Yes, if required. 
    \item \textbf{Will older versions of the dataset continue to be supported/hosted/maintained?} No, unless there is a clear productive use of an older version.  
    \item \textbf{If others want to extend/augment/build on/contribute to the dataset, is there a mechanism for them to do so?} We provide all resources for researchers to build on top of our dataset, including but not limited to regenerating the dataset, changing the logic of the dataset, adding more objects(prefabs), enhancing the physics engine for interactions, and modifying parameters of the dataset generation process. 
\end{enumerate}

%% file: main.bbl
\begin{thebibliography}{51}
\providecommand{\natexlab}[1]{#1}
\providecommand{\url}[1]{\texttt{#1}}
\expandafter\ifx\csname urlstyle\endcsname\relax
  \providecommand{\doi}[1]{doi: #1}\else
  \providecommand{\doi}{doi: \begingroup \urlstyle{rm}\Url}\fi

\bibitem[Antol et~al.(2015)Antol, Agrawal, Lu, Mitchell, Batra, Zitnick, and Parikh]{antol2015vqa}
Stanislaw Antol, Aishwarya Agrawal, Jiasen Lu, Margaret Mitchell, Dhruv Batra, C~Lawrence Zitnick, and Devi Parikh.
\newblock Vqa: Visual question answering.
\newblock In \emph{Proceedings of the IEEE international conference on computer vision}, pages 2425--2433, 2015.

\bibitem[Ashual and Wolf(2019)]{ashual2019specifying}
Oron Ashual and Lior Wolf.
\newblock Specifying object attributes and relations in interactive scene generation.
\newblock In \emph{Proceedings of the IEEE/CVF international conference on computer vision}, pages 4561--4569, 2019.

\bibitem[Borkman et~al.(2021)Borkman, Crespi, Dhakad, Ganguly, Hogins, Jhang, Kamalzadeh, Li, Leal, Parisi, et~al.]{borkman2021unity}
Steve Borkman, Adam Crespi, Saurav Dhakad, Sujoy Ganguly, Jonathan Hogins, You-Cyuan Jhang, Mohsen Kamalzadeh, Bowen Li, Steven Leal, Pete Parisi, et~al.
\newblock Unity perception: Generate synthetic data for computer vision.
\newblock \emph{arXiv preprint arXiv:2107.04259}, 2021.

\bibitem[Carreira and Zisserman(2017)]{carreira2017quo}
Joao Carreira and Andrew Zisserman.
\newblock Quo vadis, action recognition? a new model and the kinetics dataset.
\newblock In \emph{proceedings of the IEEE Conference on Computer Vision and Pattern Recognition}, pages 6299--6308, 2017.

\bibitem[Chang et~al.(2015)Chang, Funkhouser, Guibas, Hanrahan, Huang, Li, Savarese, Savva, Song, Su, et~al.]{chang2015shapenet}
Angel~X Chang, Thomas Funkhouser, Leonidas Guibas, Pat Hanrahan, Qixing Huang, Zimo Li, Silvio Savarese, Manolis Savva, Shuran Song, Hao Su, et~al.
\newblock Shapenet: An information-rich 3d model repository.
\newblock \emph{arXiv preprint arXiv:1512.03012}, 2015.

\bibitem[Chatterjee et~al.(2024)Chatterjee, Luo, Gokhale, Yang, and Baral]{chatterjee2024revision}
Agneet Chatterjee, Yiran Luo, Tejas Gokhale, Yezhou Yang, and Chitta Baral.
\newblock Revision: Rendering tools enable spatial fidelity in vision-language models.
\newblock In \emph{European Conference on Computer Vision}, pages 339--357. Springer, 2024.

\bibitem[Cho et~al.(2022)Cho, Zala, and Bansal]{DallEval}
Jaemin Cho, Abhay Zala, and Mohit Bansal.
\newblock Dall-eval: Probing the reasoning skills and social biases of text-to-image generative transformers.
\newblock \emph{arXiv preprint arXiv:2202.04053}, 2022.

\bibitem[Conwell and Ullman(2022)]{Conwell2022testing}
Colin Conwell and Tomer Ullman.
\newblock Testing relational understanding in text-guided image generation.
\newblock \emph{arXiv preprint arXiv:2208.00005}, 2022.

\bibitem[Dai et~al.(2017)Dai, Zhang, and Lin]{dai2017detecting}
Bo Dai, Yuqi Zhang, and Dahua Lin.
\newblock Detecting visual relationships with deep relational networks.
\newblock In \emph{Proceedings of the IEEE conference on computer vision and Pattern recognition}, pages 3076--3086, 2017.

\bibitem[Deng et~al.(2009{\natexlab{a}})Deng, Dong, Socher, Li, Li, and Fei-Fei]{Imagenet}
Jia Deng, Wei Dong, Richard Socher, Li-Jia Li, Kai Li, and Li Fei-Fei.
\newblock Imagenet: A large-scale hierarchical image database.
\newblock In \emph{2009 IEEE conference on computer vision and pattern recognition}, pages 248--255. Ieee, 2009{\natexlab{a}}.

\bibitem[Deng et~al.(2009{\natexlab{b}})Deng, Dong, Socher, Li, Li, and Fei-Fei]{deng2009imagenet}
Jia Deng, Wei Dong, Richard Socher, Li-Jia Li, Kai Li, and Li Fei-Fei.
\newblock Imagenet: A large-scale hierarchical image database.
\newblock In \emph{2009 IEEE conference on computer vision and pattern recognition}, pages 248--255. Ieee, 2009{\natexlab{b}}.

\bibitem[Girdhar and Ramanan(2020)]{CATER}
Rohit Girdhar and Deva Ramanan.
\newblock {CATER: A diagnostic dataset for Compositional Actions and TEmporal Reasoning}.
\newblock In \emph{ICLR}, 2020.

\bibitem[Goyal et~al.(2020)Goyal, Yang, Yang, and Deng]{Rel3d}
Ankit Goyal, Kaiyu Yang, Dawei Yang, and Jia Deng.
\newblock Rel3d: A minimally contrastive benchmark for grounding spatial relations in 3d.
\newblock \emph{Advances in Neural Information Processing Systems}, 33:\penalty0 10514--10525, 2020.

\bibitem[Goyal et~al.(2017)Goyal, Ebrahimi~Kahou, Michalski, Materzynska, Westphal, Kim, Haenel, Fruend, Yianilos, Mueller-Freitag, et~al.]{somethingsomething}
Raghav Goyal, Samira Ebrahimi~Kahou, Vincent Michalski, Joanna Materzynska, Susanne Westphal, Heuna Kim, Valentin Haenel, Ingo Fruend, Peter Yianilos, Moritz Mueller-Freitag, et~al.
\newblock The" something something" video database for learning and evaluating visual common sense.
\newblock In \emph{Proceedings of the IEEE international conference on computer vision}, pages 5842--5850, 2017.

\bibitem[Hong et~al.(2021)Hong, Yi, Tenenbaum, Torralba, and Gan]{hong2021ptr}
Yining Hong, Li Yi, Josh Tenenbaum, Antonio Torralba, and Chuang Gan.
\newblock Ptr: A benchmark for part-based conceptual, relational, and physical reasoning.
\newblock \emph{Advances in Neural Information Processing Systems}, 34:\penalty0 17427--17440, 2021.

\bibitem[Hu et~al.(2021)Hu, Shen, Wallis, Allen-Zhu, Li, Wang, Wang, and Chen]{hu2021lora}
Edward~J Hu, Yelong Shen, Phillip Wallis, Zeyuan Allen-Zhu, Yuanzhi Li, Shean Wang, Lu Wang, and Weizhu Chen.
\newblock Lora: Low-rank adaptation of large language models.
\newblock \emph{arXiv preprint arXiv:2106.09685}, 2021.

\bibitem[Hua et~al.(2022)Hua, Li, Li, Zhang, Renz, and Cohn]{actionsarerelationchains}
Hua Hua, Dongxu Li, Ruiqi Li, Peng Zhang, Jochen Renz, and Anthony Cohn.
\newblock Towards explainable action recognition by salient qualitative spatial object relation chains.
\newblock In \emph{Proceedings of the AAAI Conference on Artificial Intelligence}, pages 5710--5718, 2022.

\bibitem[Jang et~al.(2019)Jang, Song, Kim, Yu, Kim, and Kim]{jang2019video}
Yunseok Jang, Yale Song, Chris~Dongjoo Kim, Youngjae Yu, Youngjin Kim, and Gunhee Kim.
\newblock Video question answering with spatio-temporal reasoning.
\newblock \emph{International Journal of Computer Vision}, 127:\penalty0 1385--1412, 2019.

\bibitem[Ji et~al.(2020)Ji, Krishna, Fei-Fei, and Niebles]{ActionGenome}
Jingwei Ji, Ranjay Krishna, Li Fei-Fei, and Juan~Carlos Niebles.
\newblock Action genome: Actions as compositions of spatio-temporal scene graphs.
\newblock In \emph{Proceedings of the IEEE/CVF Conference on Computer Vision and Pattern Recognition}, pages 10236--10247, 2020.

\bibitem[Johnson et~al.(2017)Johnson, Hariharan, Van Der~Maaten, Fei-Fei, Lawrence~Zitnick, and Girshick]{CLEVR}
Justin Johnson, Bharath Hariharan, Laurens Van Der~Maaten, Li Fei-Fei, C Lawrence~Zitnick, and Ross Girshick.
\newblock Clevr: A diagnostic dataset for compositional language and elementary visual reasoning.
\newblock In \emph{Proceedings of the IEEE conference on computer vision and pattern recognition}, pages 2901--2910, 2017.

\bibitem[Johnson et~al.(2018)Johnson, Gupta, and Fei-Fei]{SceneGraphs}
Justin Johnson, Agrim Gupta, and Li Fei-Fei.
\newblock Image generation from scene graphs.
\newblock In \emph{Proceedings of the IEEE conference on computer vision and pattern recognition}, pages 1219--1228, 2018.

\bibitem[Kay et~al.(2017)Kay, Carreira, Simonyan, Zhang, Hillier, Vijayanarasimhan, Viola, Green, Back, Natsev, et~al.]{Kinetics}
Will Kay, Joao Carreira, Karen Simonyan, Brian Zhang, Chloe Hillier, Sudheendra Vijayanarasimhan, Fabio Viola, Tim Green, Trevor Back, Paul Natsev, et~al.
\newblock The kinetics human action video dataset.
\newblock \emph{arXiv preprint arXiv:1705.06950}, 2017.

\bibitem[Krishna et~al.(2016)Krishna, Zhu, Groth, Johnson, Hata, Kravitz, Chen, Kalantidis, Li, Shamma, Bernstein, and Fei-Fei]{VisualGenome}
Ranjay Krishna, Yuke Zhu, Oliver Groth, Justin Johnson, Kenji Hata, Joshua Kravitz, Stephanie Chen, Yannis Kalantidis, Li-Jia Li, David~A Shamma, Michael Bernstein, and Li Fei-Fei.
\newblock Visual genome: Connecting language and vision using crowdsourced dense image annotations.
\newblock 2016.

\bibitem[Krizhevsky et~al.(2009)Krizhevsky, Hinton, et~al.]{CIFAR}
Alex Krizhevsky, Geoffrey Hinton, et~al.
\newblock Learning multiple layers of features from tiny images.
\newblock 2009.

\bibitem[Li et~al.(2021)Li, Xia, Mart{\'\i}n-Mart{\'\i}n, Lingelbach, Srivastava, Shen, Vainio, Gokmen, Dharan, Jain, et~al.]{iGibson2}
Chengshu Li, Fei Xia, Roberto Mart{\'\i}n-Mart{\'\i}n, Michael Lingelbach, Sanjana Srivastava, Bokui Shen, Kent~Elliott Vainio, Cem Gokmen, Gokul Dharan, Tanish Jain, et~al.
\newblock igibson 2.0: Object-centric simulation for robot learning of everyday household tasks.
\newblock In \emph{5th Annual Conference on Robot Learning}, 2021.

\bibitem[Li et~al.(2019)Li, Yatskar, Yin, Hsieh, and Chang]{li2019visualbert}
Liunian~Harold Li, Mark Yatskar, Da Yin, Cho-Jui Hsieh, and Kai-Wei Chang.
\newblock Visualbert: A simple and performant baseline for vision and language.
\newblock \emph{arXiv preprint arXiv:1908.03557}, 2019.

\bibitem[Li et~al.(2022)Li, Wang, Zhang, Zhang, Zhao, Miao, Zhang, Tan, Wang, Wang, et~al.]{li2022end}
Mengze Li, Tianbao Wang, Haoyu Zhang, Shengyu Zhang, Zhou Zhao, Jiaxu Miao, Wenqiao Zhang, Wenming Tan, Jin Wang, Peng Wang, et~al.
\newblock End-to-end modeling via information tree for one-shot natural language spatial video grounding.
\newblock \emph{arXiv preprint arXiv:2203.08013}, 2022.

\bibitem[Litkowski and Hargraves(2021)]{TPP}
Ken Litkowski and Orin Hargraves.
\newblock The preposition project.
\newblock \emph{arXiv preprint arXiv:2104.08922}, 2021.

\bibitem[Liu et~al.(2022)Liu, Emerson, and Collier]{VSR}
Fangyu Liu, Guy Emerson, and Nigel Collier.
\newblock Visual spatial reasoning.
\newblock \emph{arXiv preprint arXiv:2205.00363}, 2022.

\bibitem[Liu et~al.(2021)Liu, Li, Du, Tenenbaum, and Torralba]{Liu2021}
Nan Liu, Shuang Li, Yilun Du, Josh Tenenbaum, and Antonio Torralba.
\newblock Learning to compose visual relations.
\newblock \emph{Advances in Neural Information Processing Systems}, 34:\penalty0 23166--23178, 2021.

\bibitem[Lu et~al.(2016)Lu, Krishna, Bernstein, and Fei-Fei]{VRD}
Cewu Lu, Ranjay Krishna, Michael Bernstein, and Li Fei-Fei.
\newblock Visual relationship detection with language priors.
\newblock In \emph{European conference on computer vision}, pages 852--869. Springer, 2016.

\bibitem[Manmadhan and Kovoor(2020)]{manmadhan2020visual}
Sruthy Manmadhan and Binsu~C Kovoor.
\newblock Visual question answering: a state-of-the-art review.
\newblock \emph{Artificial Intelligence Review}, 53:\penalty0 5705--5745, 2020.

\bibitem[Parikh and Grauman(2011)]{RelativeAttributes}
Devi Parikh and Kristen Grauman.
\newblock Relative attributes.
\newblock In \emph{2011 International Conference on Computer Vision}, pages 503--510. IEEE, 2011.

\bibitem[Radford et~al.(2021)Radford, Kim, Hallacy, Ramesh, Goh, Agarwal, Sastry, Askell, Mishkin, Clark, et~al.]{clip}
Alec Radford, Jong~Wook Kim, Chris Hallacy, Aditya Ramesh, Gabriel Goh, Sandhini Agarwal, Girish Sastry, Amanda Askell, Pamela Mishkin, Jack Clark, et~al.
\newblock Learning transferable visual models from natural language supervision.
\newblock In \emph{International conference on machine learning}, pages 8748--8763. PMLR, 2021.

\bibitem[Ramesh et~al.(2022)Ramesh, Dhariwal, Nichol, Chu, and Chen]{Ramesh2022hierarchical}
Aditya Ramesh, Prafulla Dhariwal, Alex Nichol, Casey Chu, and Mark Chen.
\newblock Hierarchical text-conditional image generation with clip latents.
\newblock \emph{arXiv preprint arXiv:2204.06125}, 2022.

\bibitem[Shang et~al.(2017)Shang, Ren, Guo, Zhang, and Chua]{shang2017video}
Xindi Shang, Tongwei Ren, Jingfan Guo, Hanwang Zhang, and Tat-Seng Chua.
\newblock Video visual relation detection.
\newblock In \emph{ACM International Conference on Multimedia}, Mountain View, CA USA, 2017.

\bibitem[Shang et~al.(2019)Shang, Di, Xiao, Cao, Yang, and Chua]{VidOR}
Xindi Shang, Donglin Di, Junbin Xiao, Yu Cao, Xun Yang, and Tat-Seng Chua.
\newblock Annotating objects and relations in user-generated videos.
\newblock In \emph{Proceedings of the 2019 on International Conference on Multimedia Retrieval}, pages 279--287. ACM, 2019.

\bibitem[Sigurdsson et~al.(2016)Sigurdsson, Varol, Wang, Farhadi, Laptev, and Gupta]{Charades}
Gunnar~A Sigurdsson, G{\"u}l Varol, Xiaolong Wang, Ali Farhadi, Ivan Laptev, and Abhinav Gupta.
\newblock Hollywood in homes: Crowdsourcing data collection for activity understanding.
\newblock In \emph{European Conference on Computer Vision}, pages 510--526. Springer, 2016.

\bibitem[Su et~al.(2021)Su, Yu, and Xu]{su2021stvgbert}
Rui Su, Qian Yu, and Dong Xu.
\newblock Stvgbert: A visual-linguistic transformer based framework for spatio-temporal video grounding.
\newblock In \emph{Proceedings of the IEEE/CVF International Conference on Computer Vision}, pages 1533--1542, 2021.

\bibitem[Thrush et~al.(2022)Thrush, Jiang, Bartolo, Singh, Williams, Kiela, and Ross]{Thrush2022winoground}
Tristan Thrush, Ryan Jiang, Max Bartolo, Amanpreet Singh, Adina Williams, Douwe Kiela, and Candace Ross.
\newblock Winoground: Probing vision and language models for visio-linguistic compositionality.
\newblock In \emph{Proceedings of the IEEE/CVF Conference on Computer Vision and Pattern Recognition}, pages 5238--5248, 2022.

\bibitem[Tran et~al.(2018)Tran, Wang, Torresani, Ray, LeCun, and Paluri]{tran2018closer}
Du Tran, Heng Wang, Lorenzo Torresani, Jamie Ray, Yann LeCun, and Manohar Paluri.
\newblock A closer look at spatiotemporal convolutions for action recognition.
\newblock In \emph{Proceedings of the IEEE conference on Computer Vision and Pattern Recognition}, pages 6450--6459, 2018.

\bibitem[Xiao et~al.(2021)Xiao, Shang, Yao, and Chua]{xiao2021next}
Junbin Xiao, Xindi Shang, Angela Yao, and Tat-Seng Chua.
\newblock Next-qa: Next phase of question-answering to explaining temporal actions.
\newblock In \emph{Proceedings of the IEEE/CVF conference on computer vision and pattern recognition}, pages 9777--9786, 2021.

\bibitem[Xu et~al.(2017)Xu, Zhu, Choy, and Fei-Fei]{VG50}
Danfei Xu, Yuke Zhu, Christopher Choy, and Li Fei-Fei.
\newblock Scene graph generation by iterative message passing.
\newblock In \emph{Computer Vision and Pattern Recognition (CVPR)}, 2017.

\bibitem[Yang et~al.(2019)Yang, Russakovsky, and Deng]{Spatialsense}
Kaiyu Yang, Olga Russakovsky, and Jia Deng.
\newblock Spatialsense: An adversarially crowdsourced benchmark for spatial relation recognition.
\newblock In \emph{Proceedings of the IEEE/CVF International Conference on Computer Vision}, pages 2051--2060, 2019.

\bibitem[Yikang et~al.()Yikang, Ouyang, and Wang]{yikangvip}
LI Yikang, Wanli Ouyang, and Xiaogang Wang.
\newblock Vip-cnn: A visual phrase reasoning convolutional neural network for visual relationship detection.

\bibitem[Yu et~al.(2019)Yu, Si, Hu, and Zhang]{yu2019review}
Yong Yu, Xiaosheng Si, Changhua Hu, and Jianxun Zhang.
\newblock A review of recurrent neural networks: Lstm cells and network architectures.
\newblock \emph{Neural computation}, 31\penalty0 (7):\penalty0 1235--1270, 2019.

\bibitem[Yusuf et~al.(2022)Yusuf, Chong, and Xianling]{yusuf2022analysis}
Abdulganiyu~Abdu Yusuf, Feng Chong, and Mao Xianling.
\newblock An analysis of graph convolutional networks and recent datasets for visual question answering.
\newblock \emph{Artificial Intelligence Review}, 55\penalty0 (8):\penalty0 6277--6300, 2022.

\bibitem[Zeng et~al.(2020)Zeng, Xu, Huang, Chen, Tan, and Gan]{zeng2020dense}
Runhao Zeng, Haoming Xu, Wenbing Huang, Peihao Chen, Mingkui Tan, and Chuang Gan.
\newblock Dense regression network for video grounding.
\newblock In \emph{Proceedings of the IEEE/CVF Conference on Computer Vision and Pattern Recognition}, pages 10287--10296, 2020.

\bibitem[Zhang et~al.(2022)Zhang, Yao, Chen, Ji, Liu, Sun, and Chua]{IETrans}
Ao Zhang, Yuan Yao, Qianyu Chen, Wei Ji, Zhiyuan Liu, Maosong Sun, and Tat-Seng Chua.
\newblock Fine-grained scene graph generation with data transfer.
\newblock \emph{ECCV}, 2022.

\bibitem[Zhang et~al.(2017{\natexlab{a}})Zhang, Kyaw, Chang, and Chua]{zhang2017visual}
Hanwang Zhang, Zawlin Kyaw, Shih-Fu Chang, and Tat-Seng Chua.
\newblock Visual translation embedding network for visual relation detection.
\newblock In \emph{Proceedings of the IEEE conference on computer vision and pattern recognition}, pages 5532--5540, 2017{\natexlab{a}}.

\bibitem[Zhang et~al.(2017{\natexlab{b}})Zhang, Kyaw, Yu, and Chang]{zhang2017ppr}
Hanwang Zhang, Zawlin Kyaw, Jinyang Yu, and Shih-Fu Chang.
\newblock Ppr-fcn: Weakly supervised visual relation detection via parallel pairwise r-fcn.
\newblock In \emph{Proceedings of the IEEE international conference on computer vision}, pages 4233--4241, 2017{\natexlab{b}}.

\end{thebibliography}
